\documentclass[11pt]{article}

\usepackage[preprint]{acl}

\usepackage{times}
\usepackage{latexsym}
\usepackage{multirow}
\usepackage{booktabs}
\usepackage{graphicx}
\usepackage{setspace}
\usepackage{amssymb}
\usepackage{enumitem}
\usepackage{pifont}
\usepackage{CJKutf8}
\usepackage[table]{xcolor}

\usepackage[most]{tcolorbox}
\usepackage{amsmath}
\usepackage[normalem]{ulem}
\usepackage[ruled,vlined,linesnumbered]{algorithm2e}

\definecolor{MyWordColor1}{RGB}{252, 245, 233}
\definecolor{MyWordColor2}{RGB}{245, 234, 212}

\usepackage{mathtools}
\usepackage[T1]{fontenc}
\usepackage[utf8]{inputenc}
\usepackage{microtype}
\usepackage{inconsolata}

\title{DynSess: Dynamic Session-Level Evaluation and Optimization Framework for Role-Playing Agents}

\author{
  \textbf{Rongsheng Zhang\textsuperscript{1,2*}},
  \textbf{Jiji Tang\textsuperscript{2*}},
  \textbf{Junnan Ren\textsuperscript{3*}},
  \textbf{Zuyi Bao\textsuperscript{2}},
\\
  \textbf{Weijie Chen\textsuperscript{2}},
  \textbf{Ruofan Hu\textsuperscript{1}},
  \textbf{Zhou Zhao\textsuperscript{1}},
  \textbf{Tangjie Lv\textsuperscript{2}},
  \textbf{Yan Zhang\textsuperscript{3\dag}}
\\
\\
  \textsuperscript{1}Zhejiang University \quad
  \textsuperscript{2}Fuxi AI Lab, NetEase Inc. \quad
  \textsuperscript{3}Xiamen University
\\
  \small{
    \texttt{\{njuzrs@163.com, tangjiji\_bupt@163.com, renjunnan@stu.xmu.edu.cn\}}
  }
}

\begin{document}
\maketitle
\renewcommand{\thefootnote}{\fnsymbol{footnote}}
\footnotetext[1]{Equal contribution.}
\footnotetext[2]{Corresponding author.}
\renewcommand{\thefootnote}{\arabic{footnote}}
\begin{abstract}
Role-playing with large language models is fundamentally a \textit{session-level} task, requiring agents to sustain character identity and interaction quality across extended multi-turn conversations. Yet existing evaluation and optimization methods remain largely turn-level, failing to capture long-horizon quality. We propose \textbf{DynSess}, a unified session-level framework for role-playing agents. \textbf{DynSess-Eval} scores complete dialogue sessions via rubrics targeting long-horizon behaviors. Leveraging its session-level rewards, we construct high-quality training trajectories through multi-turn lookahead search and train \textbf{DynSess-Character} with two complementary variants: DSPO (off-policy) and GSRPO (on-policy). Experiments show that DynSess-Eval aligns with human judgments substantially better than prior evaluators, and blind human evaluation further shows that DynSess-Character matches the strongest character model despite using substantially fewer parameters, while maintaining strong role consistency and interactive ability. Our dataset and code will be released to facilitate future research.
\end{abstract}

\section{Introduction}

\begin{figure}[!t]
  \centering
  \includegraphics[
    width=\columnwidth,
    trim=2.6cm 13.3cm 6.3cm 4.8cm,
    clip
  ]{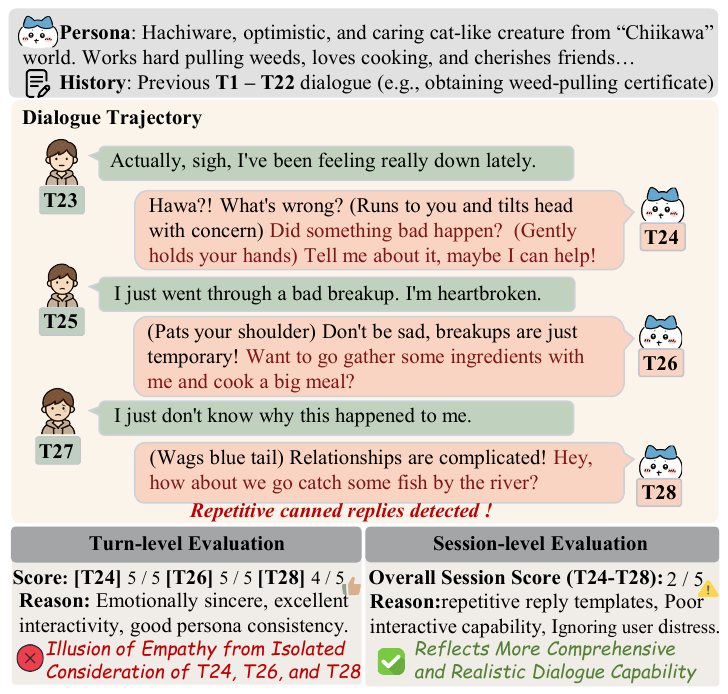}
  \caption{Turn-level vs. Session-level evaluation. An agent in a rigid response pattern scores high per turn but exhibits repetitive reply template across the session.}
  \label{fig:intro}
  \vspace{-0.6cm}
\end{figure}

Large language models (LLMs) increasingly power role-playing agents for applications such as emotional companionship, social simulation, and gaming \cite{tseng-etal-2024-two, chen2024persona}. Unlike standard instruction-following tasks, role-playing unfolds over extended, multi-turn interactions, where an agent must sustain persona fidelity, empathy, and contextual coherence across dozens of turns. Yet most existing methods for evaluating and optimizing such agents still operate at the turn level \cite{wang-etal-2024-rolellm, lu-etal-2024-large}, scoring each response in isolation. This paradigm overlooks long-horizon failure modes and consequently provides a misleading picture of an agent's true conversational competence.

As illustrated in Fig.~\ref{fig:intro}, consider an agent that ends every response with a follow-up question to a distressed user. Under turn-level evaluation, each isolated response appears highly engaging—e.g., asking "Did something happen?"—earning the agent an "excellent" score for interactivity.  However, a long-term perspective reveals a critical failure: the agent is actually trapped in a repetitive reply template. By rigidly asking questions regardless of the user's evolving emotional state, it ignores the user's actual distress and creates a mechanical experience. This exposes a fundamental flaw: turn-level metrics reward local engagement but miss global incoherence. In short, role-playing is fundamentally a session-level task—its quality emerges across turns, not within any single one.

Recent work has begun to incorporate to incorporate multi-turn history into role-playing evaluation \cite{zhou2025characterbench, xiang-etal-2025-rmtbench} and preference optimization \cite{ye-etal-2025-cpo}. While these methods condition on longer context, their signals are still computed over a \textbf{single next-turn response} against a pre-collected history, which raises two concerns. First, since the history is fixed rather than co-produced by the agent, evaluation cannot observe how the model's own earlier behavior shapes later dynamics, such as persona drift . Second, optimizing against  myopic rewards provides misaligned training signals \cite{shani2024multi, collabllm2025}, as locally preferred responses can compound into globally degraded sessions.

To address this gap, we propose \textbf{DynSess}, a unified session-level framework that couples evaluation with optimization. The evaluation component, \textbf{DynSess-Eval}, departs from prior LLM judges in two ways. First, it scores the \textbf{entire interaction trajectory} rather than individual turns, where the agent's responses and a user simulator jointly produce the dialogue. Second, it introduces a rubric explicitly designed for session-level phenomena (e.g., persona consistency across turns, adaptation to evolving user states), which cannot be defined on a single turn. Compared to standard rubric-based judges \cite{lee2025checkeval, chiang2025tract} that target static per-response attributes, this design yields evaluation signals that are both more faithful to long-horizon behavior and more robust to the score inflation that worsens as context grows.

On the optimization side, \textbf{DynSess-Character} turns session-level evaluation into trainable signals. We first construct high-quality training trajectories via a reward-driven \textbf{multi-turn lookahead search}, which selects responses by their downstream session-level reward rather than immediate plausibility, mitigating the error accumulation in Supervised Fine-tuning (\textbf{SFT}) on directly collected dialogues. Building on this, we instantiate session-level alignment under two complementary regimes: DSPO, an off-policy scheme that contrasts entire trajectories, and GSRPO, an on-policy group-relative scheme—both operating directly on session-level rewards rather than aggregating turn-level signals as in prior multi-turn Reinforcement Learning (\textbf{RL}). Together they cover off-policy and on-policy settings, letting practitioners trade off training cost against alignment quality.

Overall, our contributions are threefold:
\begin{itemize}
\item We propose \textbf{DynSess-Eval}, a rubric-anchored judge that shifts role-playing assessment from static turns to dynamic multi-turn trajectories, effectively capturing long-horizon behaviors that turn-level metrics overlook.

\item We build \textbf{DynSess-Character}, which leverages session-level evaluation for optimization through a reward-driven trajectory construction approach for enhanced SFT, together with two session-level alignment methods—\textbf{DSPO} (off-policy) and \textbf{GSRPO} (on-policy)—that jointly mitigate error accumulation across long dialogues.

\item Extensive experiments validate that \textbf{DynSess-Eval} outperforms baselines by a substantial margin in human alignment, and \textbf{DynSess-Character} achieves performance comparable to SOTA models despite using substantially fewer parameters.

\end{itemize}

\section{Related Work}

\begin{figure*}[t]
  \includegraphics[width=\textwidth,
    trim=0 6.9cm 7cm 2.1cm,
    clip
    ]{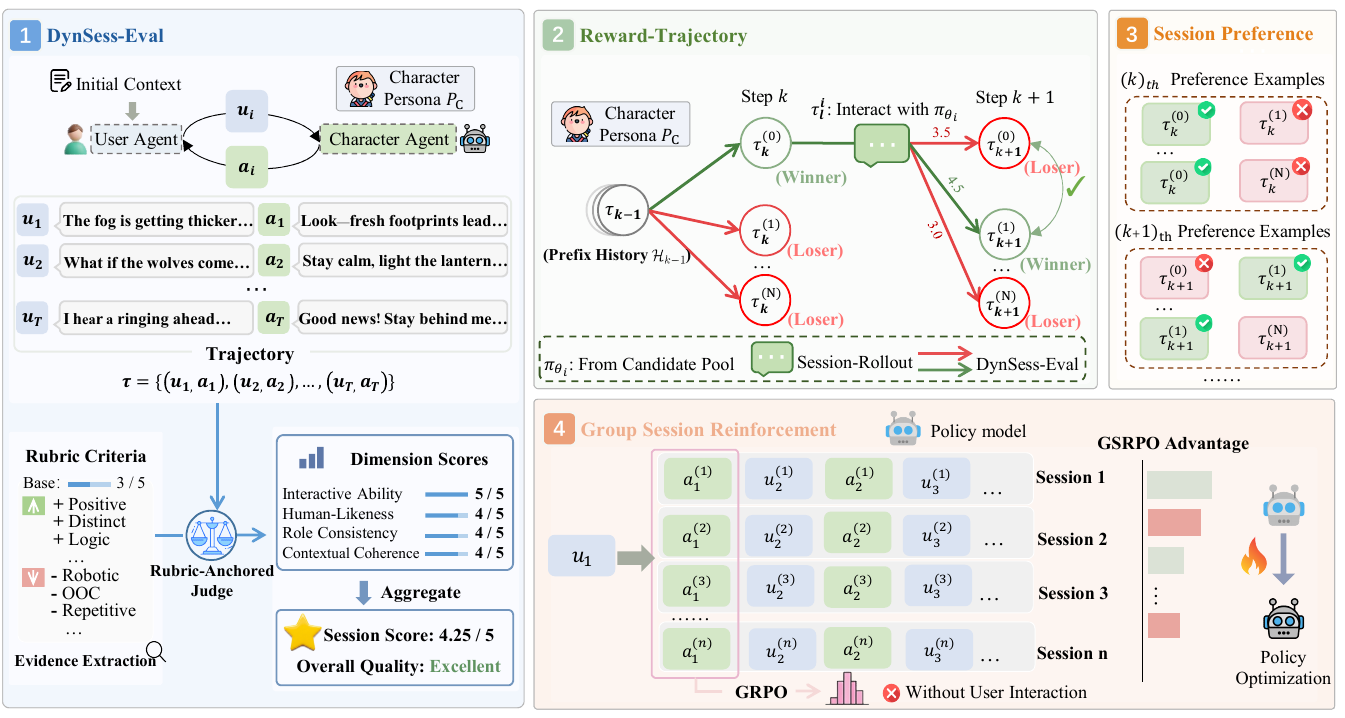}
  \caption{Overview of our framework. DynSess-Eval (left) simulates multi-turn user–agent sessions and scores them along four dimensions via a rubric-anchored judge. DynSess-Character (right) leverages these session-level rewards to construct trajectories through multi-turn lookahead search for SFT, and further aligns the agent via off-policy DSPO or on-policy GSRPO.}
  \label{fig:method}
  \vspace{-0.5cm}
\end{figure*}

\paragraph{LLM-based Role-Playing Agents.}
Early LLM-based role-playing agents mainly relied on prompting to instantiate a target character \cite{characterai}, while recent work improves persona fidelity via supervised fine-tuning on character-specific dialogues \cite{do2025aligning, wang-etal-2024-rolellm, zhou-etal-2024-characterglm, zhu-etal-2025-single} and preference-based alignment such as RLHF tailored to role-playing \cite{ye-etal-2025-cpo, fang2025act}. Despite improved response-level persona adherence, these methods remain optimized at the turn level and are poorly aligned with long-horizon quality that depends on sustained interaction and cross-turn continuity \cite{feng-etal-2025-reasoning}.

\paragraph{Evaluation of Role-Playing Agents.}
Role-playing evaluation has likewise been dominated by response-level judging. Many benchmarks adopt an LLM-as-a-Judge paradigm to score outputs against criteria such as persona fidelity, factual consistency, and stylistic matching \cite{wang-etal-2024-rolellm, li2023chatharuhi}. Representative examples include CharacterEval \cite{tu-etal-2024-charactereval} and InCharacter \cite{wang-etal-2024-incharacter}, the latter probing whether models remain in character under psychologically informed tests. Recent work has begun moving beyond static single-turn settings: CharacterBench \cite{DBLP:conf/aaai/ZhouHWBCK0XPTZZ25} uses targeted contexts to test specific capabilities, and other benchmarks introduce multi-turn interactions. Yet existing protocols still focus on isolated responses, score dialogues turn by turn, or provide only relative preference signals, offering limited evidence about long-horizon abilities such as narrative initiative, emotional progression, and contextual continuity.

\paragraph{Optimization in Multi-Turn Dialogue.}
Beyond role-playing, recent work has explored session-level optimization to capture sequential dependencies in multi-turn interaction \cite{shani2024multi, li2025beyond}. Existing methods for multi-turn RLHF \cite{gao2024regressing, zhou2024archer} and trajectory-level DPO \cite{kong2025sdpo, shi2024direct} typically target task-oriented domains, where trajectory quality can be grounded in clear terminal rewards such as task completion or solution correctness. Open-ended role-playing differs in two key ways: there is no deterministic end-state defining success, and session quality hinges on subjective properties like story flow, persona consistency, and interaction naturalness. Constructing trajectory-level preference signals for such sessions thus remains a central challenge, motivating our focus on session-level evaluation and preference optimization for role-playing agents.

\section{Methods}

In this section, we present our dynamic session-level framework for role-playing agents (illustrated in Fig.~\ref{fig:method}). As shown on the left side of the figure, DynSess-Eval serves as our evaluation component (detailed in Section \ref{sec:eval}). On the right side, DynSess-Character functions as the optimization module (detailed in Section \ref{sec:train}).

\subsection{Dynamic Session-Level Evaluation}
\label{sec:eval}

We propose Dynamic Session Evaluation (\textbf{DynSess-Eval}) to investigate the long-term conversational capabilities of role-play agents. Our key insight is that turn-level evaluation, which generates a single reply given a fixed history, lacks the longitudinal perspective to track behavioral drift and fails to account for dynamic user engagement. Therefore yields inaccurate assessments of an agent's long-term coherence.

To address this limitation, DynSess-Eval transitions from static evaluation to dynamic interaction by introducing a user simulator. As illustrated in the left panel of Fig.~\ref{fig:method}, an evaluation instance is initialized with a target character persona $P_C$ and an optional initial context $H_0$.
We first synthesize a user persona $P_U$ via a derivation module $\mathcal{M}_{\text{up}}$, parameterized by a prompt template $\mathcal{I}_{\text{up}}$ (see Appendix~\ref{app:prompt_user_persona}):
\begin{equation}
P_U \sim \mathcal{M}_{\text{up}}(\cdot \mid \mathcal{I}_{\text{up}}, P_C).
\end{equation}
Equipped with $P_C$ and $P_U$, we simulate a $T$-turn interactive session between the character agent $\pi_\theta$ and  user simulator $\pi_{\text{user}}$. At each turn $t$, both models generate responses conditioned on their respective personas and the preceding context $H_{t-1}$:
\begin{equation}
\begin{split}
u_t &\sim \pi_{\text{user}}(\cdot \mid P_U, H_{t-1}), \\
a_t &\sim \pi_{\theta}(\cdot \mid P_C, H_{t-1} \oplus u_t),
\end{split}
\end{equation}
where $u_t$ and $a_t$ denote the user utterance and agent response at turn $t$, respectively. The context is incrementally updated by appending the new utterances: $H_t = H_{t-1} \oplus (u_t, a_t)$. The resulting $T$-turn session is denoted as $\tau = H_T$.

\paragraph{Rubric-Anchored Session Scoring.}
Evaluating subjective tasks with standard LLM-as-a-judge prompts often yields inflated and unstable scores, as LLMs struggle to consistently balance multiple criteria across long, multi-turn interactions. Inspired by \cite{wei2025rocketeval}, we propose a \textbf{rubric-anchored judge} J that decouples evidence extraction from score aggregation, providing reliable and interpretable session-level signals.

Following common practices in role-playing evaluation~\cite{zhou2025characterbench}, we assess agents across four dimensions: \textit{Interactive Ability} ($D_{\text{IA}}$), \textit{Human-likeness} ($D_{\text{HL}}$), \textit{Role Consistency} ($D_{\text{RC}}$), and \textit{Contextual Coherence} ($D_{\text{CC}}$). To capture long-term dynamics,  we design dedicated \textbf{session-level criteria} $\mathcal{C}_d$ for each dimension, explicitly targeting long-horizon behaviors such as \textit{gradual persona drift}, \textit{repetitive loops}, and \textit{memory utilization}.

Specifically, given a trajectory $\tau$, the judge $J$ first uses a dimension-specific prompt $\mathcal{I}_d$ (see Appendix~\ref{app:prompt_eval}) to extract the triggered criteria:
\begin{equation}
\mathcal{E}_d(\tau) = J(\tau, \mathcal{I}_d), \quad \mathcal{E}_d(\tau) \subseteq \mathcal{C}_d.
\end{equation}
The dimension score $s_d$ is then computed by aggregating the signed weights $w_c$ of triggered criteria around a neutral baseline $b_d$, where positive weights reward desirable behaviors and negative ones penalize failure patterns:
\begin{equation}
s_d(\tau) = \mathrm{clip}\left( b_d + \sum_{\mathclap{c \in \mathcal{E}_d(\tau)}} w_c, \; s_{\min}, \; s_{\max} \right)
\label{eq:score}
\end{equation}
Both $w_c$ and $b_d$ are calibrated on a small set of human-annotated sessions to align the judge's outputs with expert ratings.
 By grounding every score in explicit evidence, our rubric-anchored judge mitigates the score inflation of standard LLM evaluators and yields a structured reward signal that directly fuels the optimization framework in §\ref{sec:train}.

\subsection{Dynamic Session-Level Optimization}
\label{sec:train}

Driven by the accuracy of our session-level evaluation in capturing long-term and dynamic behaviors, we further propose session-level optimization for character models.  As illustrated in Fig.~\ref{fig:method}, we leverage the reward signal $s_d(\tau)$ from our judge in two stages: reward-driven trajectory construction for SFT, and session-level RL via an off-policy scheme DSPO  and an on-policy scheme GSRPO.

\subsubsection{Reward-Driven Trajectory Construction}
\label{sec:stage1}

Conventional synthetic dialogues are generated myopically, turn by turn, without quality control. This local greediness causes errors to compound: a poorly chosen topic elicits perfunctory replies like "ok" or "haha", which the character obliviously elaborates upon until the session falls flat. To enable a global perspective, we leverage our session-level reward signal to introduce a \textbf{multi-turn lookahead search} that constructs trajectories segment by segment, selecting the best continuation at each step to suppress error propagation and ensure long-horizon coherence.

\begin{algorithm}[t]
\caption{Multi-turn Lookahead Search for Trajectory Construction.}
\label{alg:trajectory_construction}
\KwIn{Personas $P_C, P_U$; user simulator $\pi_{\text{user}}$; model pool $\{\pi_{\theta_i}\}_{i=1}^{N}$; session length $T$; lookahead steps $K$.}
\KwOut{All trajectory $\mathcal{H}_{K}$; rejected pairs $\mathcal{R}$.}

Initialize $\mathcal{H}_{0} \gets \emptyset$, $\mathcal{R} \gets \emptyset$\;

\For{$k=1,\ldots,K$}{
$\tau^{(i)}_{k} \gets \text{Rollout}(\mathcal{H}_{k-1}, \pi_{\text{user}}, \pi_{\theta_i}, T \mid P_C, P_U),\;\forall i \in [N]$\;

$\bar{s}_i \gets   \tfrac{1}{D}\sum_{d} s_d(\tau^{(i)}_{k}) ,\;\forall i \in [N]$\

$i^\star \gets \arg\max_{i \in [N]} \bar{s}_i$\;

$\mathcal{R} \gets \mathcal{R} \cup \bigl\{(\tau^{(i^\star)}_{k},\, \tau^{(i)}_{k}) \,\big|\, i \neq i^\star \}$\;
$\mathcal{H}_{k} \gets \mathcal{H}_{k-1} \oplus  \tau^{(i^\star)}_{k}$\;
}
\Return $\mathcal{H}_{K},\ \mathcal{R}$
\end{algorithm}

As detailed  in Algorithm~\ref{alg:trajectory_construction}, a full session $\mathcal{H}_{K}$ of total length $T \times K$ is constructed over $K$ sequential steps. At each step, a heterogeneous model pool $\{\pi_{\theta_i}\}_{i=1}^{N}$  rolls out N candidate segments in parallel, all continuing the same committed prefix $\mathcal{H}_{k-1}$.  Scoring against this shared prefix isolates segment quality from prefix variance and yields cleaner preference signals. A single search pass thus produces two aligned datasets: the committed trajectories $\mathcal{H}_{K}$ for session-level SFT, and preference pairs in $\mathcal{R}$ for the subsequent DSPO stage (§\ref{sec:stage2}).

\subsubsection{DSPO: Off-policy Session Alignment}
\label{sec:stage2}

The lookahead search in §\ref{sec:stage1} also produces a set of rejected segments $\mathcal{R}$, which provide rich contrastive signals for preference learning. A natural choice is Direct Preference Optimization (\textbf{DPO}), but its turn-level preference $(a^w \succ a^l)$ shares the same limitation as turn-level evaluation discussed in §\ref{sec:eval}—it fails to capture long-horizon behaviors.

We therefore propose Direct Session Preference Optimization (\textbf{DSPO}), which lifts the preference unit to  $T$-turn segments $(\tau^w_{k}, \tau^l_{k})$. Each pair $(\tau^w_{k}, \tau^l_{k})$ in $\mathcal{R}$ is the winner-loser selected by our judge under the same prefix $ \mathcal{H}_{k-1}$ (§\ref{sec:stage1}), as illustrated in Fig.~\ref{fig:method}. The objective is:
\begin{align}
\mathcal{L}_{\text{DSPO}} &= -\mathbb{E}_{(\tau^w_k, \tau^l_k) \sim \mathcal{R}}\bigl[\log \sigma\bigl(r_\theta(\tau^w_{k}) - r_\theta(\tau^l_{k})\bigr)\bigr], \nonumber \\
r_\theta(\tau) &= \beta \log \frac{\pi_\theta(\tau \mid \mathcal{H}_{k-1}, P_C)}{\pi_{\text{ref}}(\tau \mid \mathcal{H}_{k-1}, P_C)},
\label{eq:dspo}
\end{align}
where $\sigma$ is the logistic function, $\beta$ controls the deviation from the reference policy $\pi_{\text{ref}}$, and $r_\theta(\tau)$ is implicit segment-level preference score. Note that $\pi_\theta(\tau \mid \cdot)$ is computed only over the character tokens within $\tau$, excluding user-simulator tokens from gradient flow.

\subsubsection{GSRPO: On-policy Session RL}
\label{sec:stage3}

To complement DSPO's off-policy alignment, we introduce an on-policy stage. Standard GRPO~\cite{shao2024deepseekmath} samples M responses per user input and normalizes within-group rewards as advantages. This formulation breaks down in role-playing, where character quality emerges only across turns and single-turn rewards are noisy and weakly attributable.

We therefore propose Group Session Relative Policy Optimization (\textbf{GSRPO}), which leverages the user-agent interaction simulator from §\ref{sec:eval} to construct on-policy session groups. Specifically, instead of $M$  responses to a single prompt, we roll out $M$ full sessions $\{\tau^{m}\}_{m=1}^M$ under a shared context, and score each session with our session-level judge. The group-normalized advantage is:
\begin{equation}
\hat{A}_m = \frac{r_m - \mu_r}{\sigma_r + \varepsilon},\quad r_m = \tfrac{1}{D}\sum_{d} s_d(\tau^m),
\label{advantage}
\end{equation}
where $r_m$ is the average score across the $D$  dimensions of $s_d(\tau^m)$ from §\ref{sec:eval}, and $\mu_r, \sigma_r$ are the mean and standard deviation within the group.

Since each session $\tau^m$ interleaves character and user-simulator tokens, we broadcast $\hat{A}_m$ \emph{only} over the character token set $\mathcal{C}_m$, yielding the objective:
{\small\begin{align}
\mathcal{L}_{\text{GSRPO}} &= -\,\mathbb{E}\!\left[\tfrac{1}{M}\sum_{m=1}^{M}\tfrac{1}{|\mathcal{C}_m|}\sum_{j \in \mathcal{C}_m} \ell_{m,j}\right] + \beta\,\mathcal{R}_{\text{KL}}, \nonumber\\
\ell_{m,j} &= \min\bigl(\rho_{m,j}\hat{A}_m,\ \mathrm{clip}(\rho_{m,j}, 1{-}\epsilon, 1{+}\epsilon)\hat{A}_m\bigr),
\label{grspo}
\end{align}
}
where $\rho_{m,j} = \pi_\theta(y_j|y_{<j},P_C)/\pi_{\text{old}}(y_j|y_{<j},P_C)$ is the token-level importance ratio, and $\mathcal{R}_{\text{KL}} = \mathbb{D}_{\text{KL}}(\pi_\theta\,\|\,\pi_{\text{ref}})$ is the KL regularizer.

\section{Experiments}
\label{sec:exp-eval}
\subsection{Experiment Setup}

\renewcommand{\thefootnote}{\fnsymbol{footnote}}
\begin{table*}[t]
\caption{Comparison of DynSess-Eval against baseline evaluators in terms of alignment with human judgments. We report Rank Accuracy $\uparrow$ and Normalized Mean Absolute Error $\downarrow$ for each dimension. The best results are in \textbf{bold}.}
\vspace{-0.2cm}
\centering
\footnotesize
\setstretch{1.1}
\resizebox{\textwidth}{!}{
\begin{tabular}{lccccccccc}
\toprule[1.2pt]
\multirow{2}{*}[-0.3em]{\textbf{Eval Method}} & \multirow{2}{*}[-0.3em]{\textbf{Eval Type}} & \multicolumn{2}{c}{\textbf{Interactive Ability}} & \multicolumn{2}{c}{\textbf{Human-Likeness}} & \multicolumn{2}{c}{\textbf{Role Consistency}} & \multicolumn{2}{c}{\textbf{Contextual Coherence}} \\
\cmidrule(lr){3-4} \cmidrule(lr){5-6} \cmidrule(lr){7-8} \cmidrule(lr){9-10}
& & Rank$\uparrow$ & MAE$\downarrow$ & Rank$\uparrow$ & MAE$\downarrow$ & Rank$\uparrow$ & MAE$\downarrow$ & Rank$\uparrow$ & MAE$\downarrow$ \\
\midrule
CharacterJudge \cite{zhou2025characterbench} & Turn-Level & 0.27 & 0.51 & 0.40 & 0.45 & 0.53 & 0.41 & 0.57 & 0.54 \\
CharacterRM \cite{tu-etal-2024-charactereval} & Turn-Level & -- & -- & 0.50 & 0.37 & 0.57 & 0.37 & 0.50 & 0.51 \\
RMTBench \cite{xiang-etal-2025-rmtbench} & Pseudo-Session & 0.33 & 0.53 & 0.57 \rlap{\footnotemark[1]} & 0.39\rlap{\footnotemark[1]} & 0.50 & 0.41 & 0.53 & 0.50 \\
CharacterArena \cite{ye-etal-2025-cpo} & Trajectory Rank & 0.53 & -- & -- & -- & 0.57 & -- & 0.60 & -- \\
\midrule
\rowcolor[HTML]{EFEFEF}
\textbf{DynSess-Eval} & Session-Level & \textbf{0.83} & \textbf{0.26} & \textbf{0.77} & \textbf{0.27} & \textbf{0.67} & \textbf{0.33} & \textbf{0.73} & \textbf{0.22} \\
- w/o Rubric-Anchored  & Session-Level & 0.73 & 0.49 & 0.67 & 0.59 & 0.63 & 0.44 & 0.70 & 0.59 \\
- w/o Session-Level eval & Turn-Level & 0.20 & 0.52 & 0.53 & 0.67 & 0.43 & 0.45 & 0.63 & 0.67 \\
\bottomrule[1.2pt]
\end{tabular}
}
\label{tab:eval-model}
\vspace{-0.6cm}
\end{table*}
\renewcommand{\thefootnote}{\arabic{footnote}}

\paragraph{Datasets.}
Our experiments use a session-level role-playing dataset of 2{,}100 personas spanning diverse types (e.g., celebrities, classic literary roles, and game characters), split into a 2{,}000-persona training set and a 100-persona held-out test set, totaling 100k turns of dialogue. For each persona, we synthesize a 50-turn($T=10,K=5$) dialogue using the reward-driven lookahead search of §\ref{sec:stage1}: starting from an empty history, we perform 5 lookahead steps; at each step, 2 candidate models are sampled from a diverse pool of 5 (Gemini-3-pro, Claude Sonnet 4.6, GPT-5.4, Qwen-plus-character, and Doubao-1.5-pro-character) to roll out 10-turn segments in parallel.
The training-persona trajectories serve to optimize the character agent, whereas the eval-persona trajectories are reserved for benchmarking both character models and evaluators (detailed in the Evaluation Protocol).

\paragraph{Parameter settings.} We adopt Doubao-1.5-pro-character as the user simulator and Gemini-3-Flash as the automated evaluator for cost--efficiency trade-off, with session length set to $T=10$. Additional hyperparameters are provided in Appendix~\ref{subsec:training_details}.

\paragraph{Models and Baselines.}
\textit{Evaluator baselines} (Tab.~\ref{tab:eval-model}):
We compare against two turn-level evaluators, CharacterJudge and CharacterRM, which we adapt to session level by averaging their per-turn scores. We further compare against RMTBench (pseudo-session scoring) and CharacterArena (trajectory ranking), both used as originally implemented.
\textit{Role-playing baselines} (Tab.~\ref{tab:model-performance-human}): we cover two paradigms—general-domain LLMs (GPT-5.4, Gemini-3-pro, Claude Sonnet 4.6, DeepSeek-V3.2) and domain-specific role-playing models (Doubao-1.5-pro-character, Qwen-plus-character, MiniMax-M2-HER, Coser).

\paragraph{Evaluation Protocol.}
\textit{(1) DynSess-Eval validation} (Tab.~\ref{tab:eval-model})
 We select 8 highly discriminative personas from the test set and have three professional annotators conduct 50-turn live dialogues with each of the 5 candidate models, yielding \textbf{interactive scores grounded in real dialogue}. Each annotator then rates every model on the four DynSess-Eval dimensions, and the averaged scores form the human ground-truth ranking.
 We measure each evaluator's agreement with this ranking via Rank Accuracy (pairwise order consistency) and Normalized MAE (normalized score gap).
\textit{(2) DynSess-Character validation} (Tab.~\ref{tab:model-performance-human}, Tab.~\ref{tab:model_ablation}). For each of the 100 test personas, we uniformly sample a 0--50 turn context prefix from the synthetic trajectory and let the role-playing model produce a 10-turn continuation against our user simulator.
Each segment is scored on the four dimensions by both the annotators (averaged) and DynSess-Eval, yielding dual-evaluation signals: human ratings benchmark the role-playing models, while DynSess-Eval scores are used to guide optimization.

\begin{table}[t] \centering  \caption{Stability statistics across 9 trials per model. Variances are relatively small compared with means.}
\vspace{-0.3cm}
\label{tab:model_statistics}
\resizebox{\columnwidth}{!}{ \begin{tabular}{lccc} \toprule \textbf{Model} & \textbf{Mean ($\mu$)} & \textbf{Std. Dev ($\sigma$)} & \textbf{CV (\%)} \\ \midrule

Claude Sonnet 4.6 & 4.35 & 0.05 & 1.16 \\
GPT-5.4 & 4.29 & 0.06 & 1.39 \\
Gemini-3-pro & 4.12 & 0.04 & 0.87 \\
Doubao-1.5-pro-character & 3.96 & 0.06 & 1.45 \\
Coser-70B & 2.87 & 0.06 & 2.20 \\
\bottomrule
\end{tabular}}
\vspace{-0.3cm}
\end{table}

\subsection{Effectiveness of DynSess-Eval}

\paragraph{Alignment with Human Judgments.}

As shown in Tab.\ref{tab:eval-model}, \textbf{DynSess-Eval} achieves state-of-the-art alignment with human judgments across all four dimensions. Compared to the strongest ranking baseline CharacterArena, our method improves Rank Accuracy by 0.30 on \textit{Interactive Ability} (0.83 vs. 0.53). While compared to the strongest score-based baseline RMTBench, it reduces Normalized MAE by up to 56\% on \textit{Contextual Coherence}.

Crucially, our ablations validate both design choices. Removing rubric-anchored scoring barely affects Rank Accuracy but nearly doubles MAE (e.g., 0.22 vs 0.59 on \textit{Contextual Coherence}), showing that predefined rubrics are essential for calibrated scoring. Further degrading to turn-level evaluation (\textit{- w/o Session-Level eval}) collapses performance, with Rank Accuracy on \textit{Interactive Ability} plummeting from 0.83 to 0.20—proving turn-level metrics are blind to long-horizon dynamics and validating our session-level design.

\begin{figure}[t]
  \includegraphics[width=\columnwidth, trim = 0 23.8cm 9.1cm 0, clip]{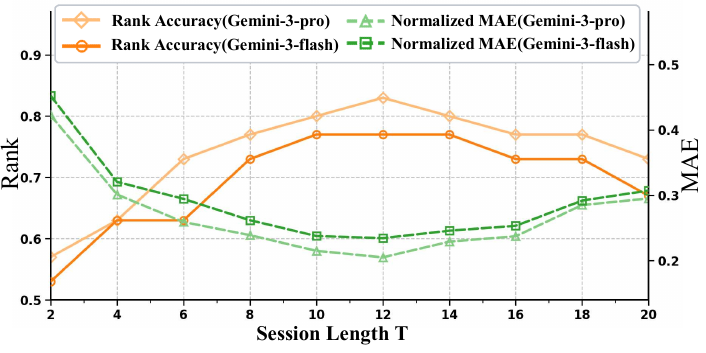}
  \vspace{-0.7cm}
  \caption{Impact of session length $T$ on evaluation. Both LLM backbones show
highly consistent trends.}
  \label{fig:analysis1}
  \vspace{-0.6cm}
\end{figure}

\paragraph{Effect of Session Length.}
We study how session length affects evaluation performance (Fig.~\ref{fig:analysis1}). Longer dialogues initially improve alignment with human by providing richer context (e.g., for role consistency), but performance plateaus and slightly drops in overly long sessions, reflecting the known degradation of LLM-as-a-judge under long contexts. Both judges achieve their best performance around $T=10$; we finanly adopt Gemini-3-Flash for its favorable cost–latency trade-off.

\paragraph{Stability Analysis.}
Given the stochasticity in LLM-based user simulation and judging, we assess the stability of DynSess-Eval. For each model, we sample three independent sessions and evaluate each three times (9 trials in total). As shown in Tab.~\ref{tab:model_statistics}, the standard deviations ($\sigma \leq 0.06$) are negligible relative to the mean scores (3.66–4.35), with coefficients of variation (CV) all below 1.7\%, indicating very low relative fluctuation. Benefiting from our carefully constructed test samples, such randomness has little impact on the overall ranking.

\begin{table*}[t]
\caption{Human evaluation results of role-playing models on four session-level dimensions, assessed via rigorous blind annotation of 10-turn continuations.  The best results are in \textbf{bold}, and the second-best are \underline{underlined}.}
\vspace{-0.3cm}
\centering
\renewcommand{\arraystretch}{1.2}
\resizebox{\textwidth}{!}{%
\begin{tabular}{l l c c c c c c}
\toprule
\multirow{2}{*}{\textbf{Domains}} & \multirow{2}{*}{\textbf{Models}} & \multirow{2}{*}{\textbf{Params}} & \multirow{2}{*}{\textbf{Average}} & \textbf{Interactive} & \textbf{Human-} & \textbf{Role} & \textbf{Contextual} \\
& & & & \textbf{Ability} & \textbf{Likeness} & \textbf{Consistency} & \textbf{Coherence} \\
\midrule
\multirow{4}{*}{General Models}
& Gemini-3-pro \cite{google2026gemini3pro} & - & 3.17 \rlap{$^\dagger$} & 3.05 \rlap{$^\dagger$} & 3.06 \rlap{$^\dagger$} & 3.36 \rlap{$^\dagger$} & 3.20 \rlap{$^\dagger$} \\
& Deepseek v3.2 \cite{liu2025deepseek} & 685B & 3.21 \rlap{$^\dagger$} & 3.11 \rlap{$^\dagger$} & 3.17 \rlap{$^\dagger$} & 3.34 \rlap{$^\dagger$} & 3.22 \\
& Claude Sonnet 4.6 \cite{anthropic2026claude45sonnet} & - & 3.24 \rlap{$^\dagger$} & 3.12 \rlap{$^\dagger$} & 3.21 \rlap{$^\dagger$} & 3.37 \rlap{$^\dagger$} & 3.27 \\
& GPT-5.4 \cite{openai2026gpt54} & - & 3.31 & 3.23 \rlap{$^\dagger$} & 3.34 & 3.38 \rlap{$^\dagger$} & 3.28 \\
\midrule
\multirow{4}{*}{Character Models}
& MiniMax-M2-HER \cite{du2026her} & 32B & 2.99 \rlap{$^\dagger$} & 2.95 \rlap{$^\dagger$} & 2.91 \rlap{$^\dagger$} & 3.13 \rlap{$^\dagger$} & 2.97 \rlap{$^\dagger$}\\
& Coser \cite{wang2025coser} & 70B & 2.96 \rlap{$^\dagger$} & 2.89 \rlap{$^\dagger$} & 2.90 \rlap{$^\dagger$} & 3.09 \rlap{$^\dagger$} & 2.98 \rlap{$^\dagger$}\\
& Qwen-plus-character \cite{yang2025qwen3} & - & 3.01 \rlap{$^\dagger$} & 2.86 \rlap{$^\dagger$} & 2.88 \rlap{$^\dagger$} & 3.22 \rlap{$^\dagger$} & 3.06 \rlap{$^\dagger$}\\
& Doubao-1.5-pro-character \cite{seed2025seed1_5thinking} & $\sim$200B\footnotemark[1] & \textbf{3.38} & \underline{3.29} \rlap{$^\dagger$} & \textbf{3.42} & \underline{3.47} \rlap{$^\dagger$} & \textbf{3.35} \\
\midrule
\multirow{2}{*}{Ours}
& \textbf{DynSess-Character-32B (DSPO)} & 32B & \underline{3.37} & 3.23 \rlap{$^\dagger$} & \underline{3.34}  & \textbf{3.56} & \underline{3.34} \\
& \textbf{DynSess-Character-32B (GSRPO)} & 32B & 3.35 & \textbf{3.39}  & 3.25 \rlap{$^\dagger$} & 3.46 \rlap{$^\dagger$} & 3.31 \\
\bottomrule
\end{tabular}%
}

\vspace{1ex}
\raggedright
\vspace{-0.1cm}
\footnotesize{$^\dagger$ significant difference with the best result (t-test, $p$-value $< 0.05$). ~200B, estimated from publicly technical reports.}
\label{tab:model-performance-human}
\vspace{-0.3cm}
\end{table*}

\begin{table*}[t]
\centering
\small
\renewcommand{\arraystretch}{1.2}
\caption{Ablation study on the Qwen3-32B backbone. We report both human evaluation (left, gold standard) and automatic evaluation (right, cross-validation).  All preference and RL variants (DPO, GRPO, DSPO, GSRPO) are trained on top of the Reward-Driven Trajector checkpoint. }
\vspace{-0.3cm}
\label{tab:model_ablation}
\resizebox{\textwidth}{!}{%
\begin{tabular}{l c | c c c c c | c c c c c}
\toprule
\multirow{2}{*}{\textbf{Model}} & \multirow{2}{*}{\textbf{Reward Level}} & \multicolumn{5}{c|}{\textbf{Human Evaluation}} & \multicolumn{5}{c}{\textbf{Auto Evaluation}} \\
\cmidrule(lr){3-7} \cmidrule(lr){8-12}
& & \textbf{Avg.} & \textbf{Inter.} & \textbf{Human-L.} & \textbf{Role} & \textbf{Context} & \textbf{Avg.} & \textbf{Inter.} & \textbf{Human-L.} & \textbf{Role} & \textbf{Context} \\
\midrule
Claude Sonnet 4.6 & -- & 3.24 & 3.12 & 3.21 & 3.37 & 3.27 & \underline{4.35} & 3.76 & \textbf{4.55} & \textbf{4.88} & 4.24 \\
GPT-5.4 & -- & 3.31 & \underline{3.23} & \textbf{3.34} & 3.38 & 3.28 & 4.29 & 3.80 & 4.27 & \textbf{4.88} & \underline{4.29} \\
\midrule

SFT (baseline) & -- & 3.22 & 3.12 & 3.16 & 3.41 & 3.17 & 4.02 & 3.49 & 3.96 & 4.70 & 3.90 \\
w/\{Reward-Driven Trajectory \} & -- & 3.32 & 3.21 & 3.31 & 3.52 & 3.26 & 4.16 & 3.74 & 4.13 & 4.80 & 4.01 \\
\midrule
w/\{DPO\} & Turn & 3.22 & 3.19 & 3.22 & 3.30 & 3.16 & 4.17 & 3.73 & 4.05 & 4.80 & 4.11 \\
w/\{GRPO\} & Turn & 3.34 & 3.20 & 3.32 & \underline{3.55} & \underline{3.32} & 4.30 & \underline{3.93} & 4.38 & 4.84 & 4.06 \\
\midrule
w/\{DSPO\} & Session & \textbf{3.37} & 3.23 & \textbf{3.34} & \textbf{3.56} & \textbf{3.34} & 4.28 & 3.88 & 4.32 & 4.83 & 4.09 \\
w/\{GSRPO\} & Session & \underline{3.35} & \textbf{3.39} & 3.25 & 3.46 & 3.31 &  \textbf{4.46} & \textbf{4.15} & \underline{4.50} & 4.87 & \textbf{4.31} \\
\bottomrule
\end{tabular}%
}
\vspace{-0.6cm}
\end{table*}

\subsection{Performance of DynSess-Character}
\paragraph{Main Results.}
Performance comparisons with existing baselines are illustrated in Tab.~\ref{tab:model-performance-human}. We further conduct paired t-tests against the best-performing model in each column, with significant gaps (p < 0.05) marked.
Our DSPO variant achieves the highest Role Consistency score (3.56), and GSRPO leads on Interactive Ability (3.39), both with statistically significant margins over all baselines. Both variants confirm that  session-level reinforcement learning effectively encourages proactive narrative driving rather than passive responding. Notably, with only 32B parameters, DSPO reaches an average score (3.37) comparable to the leading proprietary model Doubao-1.5-pro-character (3.38, 200B), yielding a nearly 6 times parameter efficiency. Absolute margins between top models are narrow (typically 0.05--0.15), reflecting the well-known central tendency bias in Likert-scale evaluation. Nevertheless, our annotators reach a strong Spearman correlation of over \textbf{0.80} (Appendix~\ref{app:iaa}), confirming reliable rankings.

\begin{figure*}[t]
\centering
\includegraphics[width=0.95\textwidth, trim = 0 19cm 5.5cm 0, clip]{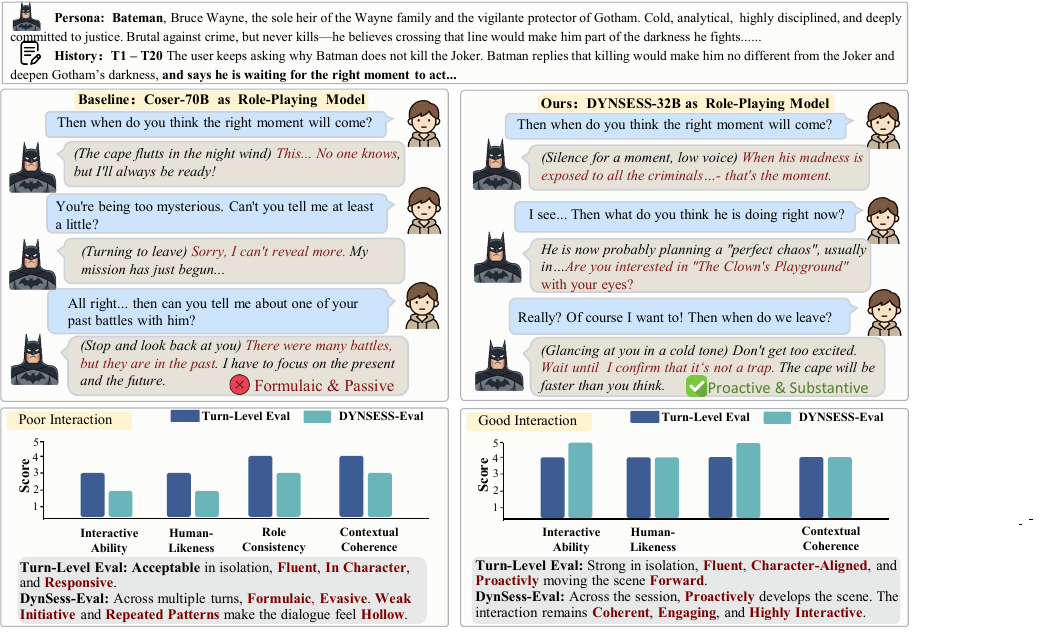}
\caption{A case study on a Batman role-playing session. Top: Coser-70B baseline vs.\ DynSess-Character-32B. Bottom: turn-level vs.\ DynSess-Eval (session-level) evaluators.}
\label{fig:case}
\vspace{-0.5cm}
\end{figure*}

\paragraph{Component Ablation.}
Ablation results on the Qwen3-32B backbone are reported in Table~\ref{tab:model_ablation}, using human evaluation as the  main metric and automatic evaluation for complementary analysis.
First, incorporating our \textit{Reward-Driven Trajectory} construction strategy consistently improves the SFT baseline from 3.22 to 3.32, confirming that high-quality session-level data directly enhances model performance.
Second, session-level optimization consistently outperforms turn-level counterparts across both off-policy (DSPO vs DPO) and  on-policy RL (GSRPO vs GRPO). Specifically, DSPO achieves the highest overall Human score (3.37), while GSRPO further leads Interactive Ability (3.39)—the highest among all variants. These gains demonstrate that session-level reward modeling effectively captures long-horizon dependencies, substantially strengthening the agent's ability to sustain coherent and persona-consistent dialogues.

\paragraph{Discrepancy Between Auto and Human Eval.}
An intriguing phenomenon emerges from Tab.~\ref{tab:model_ablation}: automatic and human evaluations frequently disagree, even when the auto-evaluator is calibrated against human judgments. We highlight three representative cases.
(i) \textbf{Residual misalignment in subjective dimensions}. Despite DynSess-Eval being rigorously calibrated to human preferences, GSRPO obtains a notably higher Auto score on Human-Likeness (4.50) yet only a moderate Human score (3.25), suggesting that highly subjective dimensions remain hard to fully align.
(ii) \textbf{Self-preference bias of LLM-as-a-judge}. Closed-source models such as Claude Sonnet 4.6 dominate Auto evaluation (Avg. 4.35) but rank poorly under Human evaluation (Avg. 3.24), reflecting a systematic preference of LLM judges toward outputs stylistically similar to their own.
(iii) \textbf{On-policy training overfits evaluator}. While on-policy RL is conventionally regarded as the stronger paradigm, on-policy GSRPO surpasses off-policy DSPO on Auto (4.46 vs.\ 4.28) but underperforms on Human (3.35 vs.\ 3.37), indicating that on-policy methods are more prone to fitting evaluator-specific artifacts.
Collectively, these observations motivate our use of huma evaluation as the gold standard.

\subsection{Case Study}

Fig.~\ref{fig:case} compares the turn-level baseline and DynSess-32B on a multi-turn Batman dialogue.  Both start fluent, but diverge over long horizons. Turn-level model drifts into an evasive, user-mirroring tone and eventually degrades into a generic AI assistant, while DynSess-32B sustains Batman's signature traits (terse, brooding, morally complex) and proactively advances the scene with new conflicts and inner motivations.

The lower half of Fig.~\ref{fig:case} further exposes the limitation of turn-level evaluators. In Case 1, a turn-level judge rates the baseline highly from isolated early turns, while DynSess-Eval penalizes its templated drift in later turns. In Case 2, the session-level view further enables DynSess-Eval to capture how DynSess-32B adapts to the user's evolving emotions—a long-horizon trait turn-level metrics cannot capture. Together, these cases confirm that session-level modeling is indispensable for both the optimization and evaluation of role-playing agents.

\section{Conclusion}

In this work, we study role-playing as a session-level task rather than isolated turns, instantiating this view through a unified framework. DynSess-Eval provides session-level evaluation that improves Rank Accuracy and reduces Normalized MAE over the strongest baselines. Built on its session-level rewards, DSPO and GSRPO extend alignment from single turns to full trajectories. Under rigorous session-continuation evaluation, DynSess-Character-32B matches a ~200B proprietary model overall while leading in proactive, long-horizon dimensions. Our joint auto–human analysis reveals residual evaluator biases, reinforcing the need for human-in-the-loop judgment. We hope this establishes session-level modeling as a foundation for consistent, adaptive, and human-aligned role-playing agents.

\section{Limitations}
Our work has several limitations worth noting.
First, DynSess-Eval relies on a single LLM (Gemini-3-Flash) as the backbone judge. Although its rubric-anchored design substantially improves alignment with human judgments, the residual auto--human discrepancies observed in Section~\ref{sec:exp-eval} suggest that no single judge model can fully substitute for human evaluation. Second, our experiments cap session length at 50 turns. Although Section~\ref{sec:exp-eval} shows that evaluation quality plateaus and slightly degrades in excessively long contexts, real-world role-playing deployments may involve substantially longer interactions, and extending our framework to such regimes requires further investigation.

\section{Ethics Statement}
Our work involves human annotators and LLM-generated dialogues, raising several ethical considerations. All human annotators participated voluntarily and were compensated fairly above local standards. Annotators were informed of the research purpose and their right to withdraw at any time. The character personas used in our experiments are drawn from publicly available sources (e.g., fictional characters, public figures). We do not generate personas that impersonate private individuals or promote harmful stereotypes. While role-playing agents can serve beneficial purposes such as entertainment and social companionship, we acknowledge the potential for misuse, including generating misleading or emotionally manipulative content. We advocate for deploying such systems with appropriate safeguards.

\bibliography{custom}

@inproceedings{wang-etal-2024-rolellm,
    title = "{R}ole{LLM}: Benchmarking, Eliciting, and Enhancing Role-Playing Abilities of Large Language Models",
    author = "Wang, Noah  and
      Peng, Z.y.  and
      Que, Haoran  and
      Liu, Jiaheng  and
      Zhou, Wangchunshu  and
      Wu, Yuhan  and
      Guo, Hongcheng  and
      Gan, Ruitong  and
      Ni, Zehao  and
      Yang, Jian  and
      Zhang, Man  and
      Zhang, Zhaoxiang  and
      Ouyang, Wanli  and
      Xu, Ke  and
      Huang, Wenhao  and
      Fu, Jie  and
      Peng, Junran",
    editor = "Ku, Lun-Wei  and
      Martins, Andre  and
      Srikumar, Vivek",
    booktitle = "Findings of the Association for Computational Linguistics: ACL 2024",
    month = aug,
    year = "2024",
    address = "Bangkok, Thailand",
    publisher = "Association for Computational Linguistics",
    url = "https://aclanthology.org/2024.findings-acl.878/",
    doi = "10.18653/v1/2024.findings-acl.878",
    pages = "14743--14777"
}

@inproceedings{zhou-etal-2024-characterglm,
    title = "{C}haracter{GLM}: Customizing Social Characters with Large Language Models",
    author = "Zhou, Jinfeng  and
      Chen, Zhuang  and
      Wan, Dazhen  and
      Wen, Bosi  and
      Song, Yi  and
      Yu, Jifan  and
      Huang, Yongkang  and
      Ke, Pei  and
      Bi, Guanqun  and
      Peng, Libiao  and
      Yang, JiaMing  and
      Xiao, Xiyao  and
      Sabour, Sahand  and
      Zhang, Xiaohan  and
      Hou, Wenjing  and
      Zhang, Yijia  and
      Dong, Yuxiao  and
      Wang, Hongning  and
      Tang, Jie  and
      Huang, Minlie",
    editor = "Dernoncourt, Franck  and
      Preo{\c{t}}iuc-Pietro, Daniel  and
      Shimorina, Anastasia",
    booktitle = "Proceedings of the 2024 Conference on Empirical Methods in Natural Language Processing: Industry Track",
    month = nov,
    year = "2024",
    address = "Miami, Florida, US",
    publisher = "Association for Computational Linguistics",
    url = "https://aclanthology.org/2024.emnlp-industry.107/",
    doi = "10.18653/v1/2024.emnlp-industry.107",
    pages = "1457--1476"
}

@article{fang2025act,
  title={Act-Adaptive Margin: Dynamically Calibrating Reward Models for Subjective Ambiguity},
  author={Fang, Feiteng and Chen, Dingwei and Huang, Xiang and Lin, Ting-En and Wu, Yuchuan and Liu, Xiong and Ye, Xinge and Liu, Ziqiang and Zhang, Haonan and Zhu, Liang and others},
  journal={arXiv preprint arXiv:2505.23923},
  year={2025}
}

@inproceedings{do2025aligning,
  title={Aligning large language models with human opinions through persona selection and value--belief--norm reasoning},
  author={Do, Xuan Long and Kawaguchi, Kenji and Kan, Min-Yen and Chen, Nancy},
  booktitle={Proceedings of the 31st International Conference on Computational Linguistics},
  pages={2526--2547},
  year={2025}
}

@inproceedings{zhu-etal-2025-single,
    title = "Single {LLM}, Multiple Roles: A Unified Retrieval-Augmented Generation Framework Using Role-Specific Token Optimization",
    author = "Zhu, Yutao  and
      Jin, Jiajie  and
      Qian, Hongjin  and
      Liu, Zheng  and
      Dou, Zhicheng  and
      Wen, Ji-Rong",
    editor = "Christodoulopoulos, Christos  and
      Chakraborty, Tanmoy  and
      Rose, Carolyn  and
      Peng, Violet",
    booktitle = "Proceedings of the 2025 Conference on Empirical Methods in Natural Language Processing",
    month = nov,
    year = "2025",
    address = "Suzhou, China",
    publisher = "Association for Computational Linguistics",
    url = "https://aclanthology.org/2025.emnlp-main.243/",
    doi = "10.18653/v1/2025.emnlp-main.243",
    pages = "4837--4856",
    ISBN = "979-8-89176-332-6"
}

@inproceedings{feng-etal-2025-reasoning,
    title = "Reasoning Does Not Necessarily Improve Role-Playing Ability",
    author = "Feng, Xiachong  and
      Dou, Longxu  and
      Kong, Lingpeng",
    editor = "Che, Wanxiang  and
      Nabende, Joyce  and
      Shutova, Ekaterina  and
      Pilehvar, Mohammad Taher",
    booktitle = "Findings of the Association for Computational Linguistics: ACL 2025",
    month = jul,
    year = "2025",
    address = "Vienna, Austria",
    publisher = "Association for Computational Linguistics",
    url = "https://aclanthology.org/2025.findings-acl.537/",
    doi = "10.18653/v1/2025.findings-acl.537",
    pages = "10301--10314",
    ISBN = "979-8-89176-256-5"
}

@misc{characterai,
  author       = {{Character.AI Inc.}},
  title        = {Character.AI},
  howpublished = {\url{https://character.ai/}},
  note         = {Accessed: 2024-05-20},
  year         = {2024}
}

@inproceedings{wang-etal-2024-incharacter,
    title = "{I}n{C}haracter: Evaluating Personality Fidelity in Role-Playing Agents through Psychological Interviews",
    author = "Wang, Xintao  and
      Xiao, Yunze  and
      Huang, Jen-tse  and
      Yuan, Siyu  and
      Xu, Rui  and
      Guo, Haoran  and
      Tu, Quan  and
      Fei, Yaying  and
      Leng, Ziang  and
      Wang, Wei  and
      Chen, Jiangjie  and
      Li, Cheng  and
      Xiao, Yanghua",
    editor = "Ku, Lun-Wei  and
      Martins, Andre  and
      Srikumar, Vivek",
    booktitle = "Proceedings of the 62nd Annual Meeting of the Association for Computational Linguistics (Volume 1: Long Papers)",
    month = aug,
    year = "2024",
    address = "Bangkok, Thailand",
    publisher = "Association for Computational Linguistics",
    url = "https://aclanthology.org/2024.acl-long.102/",
    doi = "10.18653/v1/2024.acl-long.102",
    pages = "1840--1873"
}

@inproceedings{tu-etal-2024-charactereval,
    title = "{C}haracter{E}val: A {C}hinese Benchmark for Role-Playing Conversational Agent Evaluation",
    author = "Tu, Quan  and
      Fan, Shilong  and
      Tian, Zihang  and
      Shen, Tianhao  and
      Shang, Shuo  and
      Gao, Xin  and
      Yan, Rui",
    editor = "Ku, Lun-Wei  and
      Martins, Andre  and
      Srikumar, Vivek",
    booktitle = "Proceedings of the 62nd Annual Meeting of the Association for Computational Linguistics (Volume 1: Long Papers)",
    month = aug,
    year = "2024",
    address = "Bangkok, Thailand",
    publisher = "Association for Computational Linguistics",
    url = "https://aclanthology.org/2024.acl-long.638/",
    doi = "10.18653/v1/2024.acl-long.638",
    pages = "11836--11850"
}

@article{li2023chatharuhi,
  title={Chatharuhi: Reviving anime character in reality via large language model},
  author={Li, Cheng and Leng, Ziang and Yan, Chenxi and Shen, Junyi and Wang, Hao and Mi, Weishi and Fei, Yaying and Feng, Xiaoyang and Yan, Song and Wang, HaoSheng and others},
  journal={arXiv preprint arXiv:2308.09597},
  year={2023}
}

@inproceedings{DBLP:conf/aaai/ZhouHWBCK0XPTZZ25,
  author       = {Jinfeng Zhou and
                  Yongkang Huang and
                  Bosi Wen and
                  Guanqun Bi and
                  Yuxuan Chen and
                  Pei Ke and
                  Zhuang Chen and
                  Xiyao Xiao and
                  Libiao Peng and
                  Kuntian Tang and
                  Rongsheng Zhang and
                  Le Zhang and
                  Tangjie Lv and
                  Zhipeng Hu and
                  Hongning Wang and
                  Minlie Huang},
  editor       = {Toby Walsh and
                  Julie Shah and
                  Zico Kolter},
  title        = {CharacterBench: Benchmarking Character Customization of Large Language
                  Models},
  booktitle    = {AAAI-25, Sponsored by the Association for the Advancement of Artificial
                  Intelligence, February 25 - March 4, 2025, Philadelphia, PA, {USA}},
  pages        = {26101--26110},
  publisher    = {{AAAI} Press},
  year         = {2025},
  url          = {https://doi.org/10.1609/aaai.v39i24.34806},
  doi          = {10.1609/AAAI.V39I24.34806},
  bibsource    = {dblp computer science bibliography, https://dblp.org}
}

@inproceedings{zhou2025characterbench,
  title={CharacterBench: benchmarking character customization of large language models},
  author={Zhou, Jinfeng and Huang, Yongkang and Wen, Bosi and Bi, Guanqun and Chen, Yuxuan and Ke, Pei and Chen, Zhuang and Xiao, Xiyao and Peng, Libiao and Tang, Kuntian and others},
  booktitle={Proceedings of the AAAI Conference on Artificial Intelligence},
  volume={39},
  number={24},
  pages={26101--26110},
  year={2025}
}

@article{shao2024deepseekmath,
  title={Deepseekmath: Pushing the limits of mathematical reasoning in open language models},
  author={Shao, Zhihong and Wang, Peiyi and Zhu, Qihao and Xu, Runxin and Song, Junxiao and Bi, Xiao and Zhang, Haowei and Zhang, Mingchuan and Li, YK and Wu, Yang and others},
  journal={arXiv preprint arXiv:2402.03300},
  year={2024}
}

@article{shani2024multi,
  title={Multi-turn reinforcement learning with preference human feedback},
  author={Shani, Lior and Rosenberg, Aviv and Cassel, Asaf and Lang, Oran and Calandriello, Daniele and Zipori, Avital and Noga, Hila and Keller, Orgad and Piot, Bilal and Szpektor, Idan and others},
  journal={Advances in Neural Information Processing Systems},
  volume={37},
  pages={118953--118993},
  year={2024}
}

@article{li2025beyond,
  title={Beyond single-turn: A survey on multi-turn interactions with large language models},
  author={Li, Yubo and Shen, Xiaobin and Yao, Xinyu and Ding, Xueying and Miao, Yidi and Krishnan, Ramayya and Padman, Rema},
  journal={arXiv preprint arXiv:2504.04717},
  year={2025}
}

@article{gao2024regressing,
  title={Regressing the relative future: Efficient policy optimization for multi-turn rlhf},
  author={Gao, Zhaolin and Zhan, Wenhao and Chang, Jonathan D and Swamy, Gokul and Brantley, Kiant{\'e} and Lee, Jason D and Sun, Wen},
  journal={arXiv preprint arXiv:2410.04612},
  year={2024}
}

@article{zhou2024archer,
  title={Archer: Training language model agents via hierarchical multi-turn rl},
  author={Zhou, Yifei and Zanette, Andrea and Pan, Jiayi and Levine, Sergey and Kumar, Aviral},
  journal={arXiv preprint arXiv:2402.19446},
  year={2024}
}

@inproceedings{kong2025sdpo,
  title={Sdpo: Segment-level direct preference optimization for social agents},
  author={Kong, Aobo and Ma, Wentao and Zhao, Shiwan and Li, Yongbin and Wu, Yuchuan and Wang, Ke and Liu, Xiaoqian and Li, Qicheng and Qin, Yong and Huang, Fei},
  booktitle={Proceedings of the 63rd Annual Meeting of the Association for Computational Linguistics (Volume 1: Long Papers)},
  pages={12409--12423},
  year={2025}
}

@inproceedings{shi2024direct,
  title={Direct multi-turn preference optimization for language agents},
  author={Shi, Wentao and Yuan, Mengqi and Wu, Junkang and Wang, Qifan and Feng, Fuli},
  booktitle={Proceedings of the 2024 Conference on Empirical Methods in Natural Language Processing},
  pages={2312--2324},
  year={2024}
}

@inproceedings{tseng-etal-2024-two,
    title = "Two Tales of Persona in {LLM}s: A Survey of Role-Playing and Personalization",
    author = "Tseng, Yu-Min  and
      Huang, Yu-Chao  and
      Hsiao, Teng-Yun  and
      Chen, Wei-Lin  and
      Huang, Chao-Wei  and
      Meng, Yu  and
      Chen, Yun-Nung",
    editor = "Al-Onaizan, Yaser  and
      Bansal, Mohit  and
      Chen, Yun-Nung",
    booktitle = "Findings of the Association for Computational Linguistics: EMNLP 2024",
    month = nov,
    year = "2024",
    address = "Miami, Florida, USA",
    publisher = "Association for Computational Linguistics",
    url = "https://aclanthology.org/2024.findings-emnlp.969/",
    doi = "10.18653/v1/2024.findings-emnlp.969",
    pages = "16612--16631"
}

@article{chen2024persona,
  title={From persona to personalization: A survey on role-playing language agents},
  author={Chen, Jiangjie and Wang, Xintao and Xu, Rui and Yuan, Siyu and Zhang, Yikai and Shi, Wei and Xie, Jian and Li, Shuang and Yang, Ruihan and Zhu, Tinghui and others},
  journal={arXiv preprint arXiv:2404.18231},
  year={2024}
}

@inproceedings{lu-etal-2024-large,
    title = "Large Language Models are Superpositions of All Characters: Attaining Arbitrary Role-play via Self-Alignment",
    author = "Lu, Keming  and
      Yu, Bowen  and
      Zhou, Chang  and
      Zhou, Jingren",
    editor = "Ku, Lun-Wei  and
      Martins, Andre  and
      Srikumar, Vivek",
    booktitle = "Proceedings of the 62nd Annual Meeting of the Association for Computational Linguistics (Volume 1: Long Papers)",
    month = aug,
    year = "2024",
    address = "Bangkok, Thailand",
    publisher = "Association for Computational Linguistics",
    url = "https://aclanthology.org/2024.acl-long.423/",
    doi = "10.18653/v1/2024.acl-long.423",
    pages = "7828--7840"
}

@inproceedings{xiang-etal-2025-rmtbench,
    title = "{RMTB}ench: Benchmarking {LLM}s Through Multi-Turn User-Centric Role-Playing",
    author = "Xiang, Hao  and
      Tang, Tianyi  and
      Su, Yang  and
      Yu, Bowen  and
      Yang, An  and
      Huang, Fei  and
      Zhang, Yichang  and
      Lu, Yaojie  and
      Lin, Hongyu  and
      Han, Xianpei  and
      Zhou, Jingren  and
      Lin, Junyang  and
      Sun, Le",
    editor = "Christodoulopoulos, Christos  and
      Chakraborty, Tanmoy  and
      Rose, Carolyn  and
      Peng, Violet",
    booktitle = "Findings of the Association for Computational Linguistics: EMNLP 2025",
    month = nov,
    year = "2025",
    address = "Suzhou, China",
    publisher = "Association for Computational Linguistics",
    url = "https://aclanthology.org/2025.findings-emnlp.730/",
    doi = "10.18653/v1/2025.findings-emnlp.730",
    pages = "13555--13571",
    ISBN = "979-8-89176-335-7"
}

@inproceedings{ye-etal-2025-cpo,
    title = "{CPO}: Addressing Reward Ambiguity in Role-playing Dialogue via Comparative Policy Optimization",
    author = "Ye, Jing  and
      Wang, Rui  and
      Wu, Yuchuan  and
      Ma, Victor  and
      Fang, Feiteng  and
      Huang, Fei  and
      Li, Yongbin",
    editor = "Christodoulopoulos, Christos  and
      Chakraborty, Tanmoy  and
      Rose, Carolyn  and
      Peng, Violet",
    booktitle = "Findings of the Association for Computational Linguistics: EMNLP 2025",
    month = nov,
    year = "2025",
    address = "Suzhou, China",
    publisher = "Association for Computational Linguistics",
    url = "https://aclanthology.org/2025.findings-emnlp.18/",
    doi = "10.18653/v1/2025.findings-emnlp.18",
    pages = "297--323",
    ISBN = "979-8-89176-335-7"
}

@article{liu2025deepseek,
  title={Deepseek-v3. 2: Pushing the frontier of open large language models},
  author={Liu, Aixin and Mei, Aoxue and Lin, Bangcai and Xue, Bing and Wang, Bingxuan and Xu, Bingzheng and Wu, Bochao and Zhang, Bowei and Lin, Chaofan and Dong, Chen and others},
  journal={arXiv preprint arXiv:2512.02556},
  year={2025}
}

@misc{openai2026gpt54,
  author = {{OpenAI}},
  title = {GPT-5.4},
  year = {2026},
  howpublished = {\url{https://openai.com/}},
  note = {Large language model. Accessed: 2026-03-23}
}

@misc{google2026gemini3pro,
  author = {{Google DeepMind}},
  title = {Gemini 3 Pro},
  year = {2026},
  howpublished = {\url{https://deepmind.google/technologies/gemini/}},
  note = {Large language model. Accessed: 2026-03-23}
}

@misc{anthropic2026claude45sonnet,
  author = {{Anthropic}},
  title = {Claude 4.5 Sonnet},
  year = {2026},
  howpublished = {\url{https://www.anthropic.com/}},
  note = {Large language model. Accessed: 2026-03-23}
}

@inproceedings{wang2025coser,
  title={Coser: Coordinating llm-based persona simulation of established roles},
  author={Wang, Xintao and Wang, Heng and Zhang, Yifei and Yuan, Xinfeng and Xu, Rui and Huang, Jen-tse and Yuan, Siyu and Guo, Haoran and Chen, Jiangjie and Zhou, Shuchang and others},
  booktitle={Forty-second International Conference on Machine Learning},
  year={2025}
}

@article{du2026her,
  title={HER: Human-like Reasoning and Reinforcement Learning for LLM Role-playing},
  author={Du, Chengyu and Wang, Xintao and Chen, Aili and Li, Weiyuan and Xu, Rui and Liu, Junteng and Huang, Zishan and Tian, Rong and Sun, Zijun and Li, Yuhao and others},
  journal={arXiv preprint arXiv:2601.21459},
  year={2026}
}

@article{seed2025seed1_5thinking,
  title   = {Seed1.5-Thinking: Advancing Superb Reasoning Models with Reinforcement Learning},
  author  = {ByteDance Seed},
  journal = {arXiv preprint arXiv:2504.13914},
  year    = {2025}
}

@article{yang2025qwen3,
  title={Qwen3 technical report},
  author={Yang, An and Li, Anfeng and Yang, Baosong and Zhang, Beichen and Hui, Binyuan and Zheng, Bo and Yu, Bowen and Gao, Chang and Huang, Chengen and Lv, Chenxu and others},
  journal={arXiv preprint arXiv:2505.09388},
  year={2025}
}

@inproceedings{zheng2024llamafactory,
  title={Llamafactory: Unified efficient fine-tuning of 100+ language models},
  author={Zheng, Yaowei and Zhang, Richong and Zhang, Junhao and Ye, Yanhan and Luo, Zheyan},
  booktitle={Proceedings of the 62nd annual meeting of the association for computational linguistics (volume 3: system demonstrations)},
  pages={400--410},
  year={2024}
}

@inproceedings{collabllm2025,
    title={CollabLLM: From Passive Responders to Active Collaborators},
    author={Shirley Wu and Michel Galley and Baolin Peng and Hao Cheng and 
            Gavin Li and Yao Dou and Weixin Cai and James Zou and 
            Jure Leskovec and Jianfeng Gao},
    booktitle={International Conference on Machine Learning (ICML)},
    year={2025}
}

@inproceedings{lee2025checkeval,
  title={Checkeval: A reliable llm-as-a-judge framework for evaluating text generation using checklists},
  author={Lee, Yukyung and Kim, Joonghoon and Kim, Jaehee and Cho, Hyowon and Kang, Jaewook and Kang, Pilsung and Kim, Najoung},
  booktitle={Proceedings of the 2025 Conference on Empirical Methods in Natural Language Processing},
  pages={15782--15809},
  year={2025}
}

@inproceedings{chiang2025tract,
  title={Tract: Regression-aware fine-tuning meets chain-of-thought reasoning for llm-as-a-judge},
  author={Chiang, Cheng-Han and Lee, Hung-yi and Lukasik, Michal},
  booktitle={Proceedings of the 63rd Annual Meeting of the Association for Computational Linguistics (Volume 1: Long Papers)},
  pages={2934--2952},
  year={2025}
}

@inproceedings{wei2025rocketeval,
title={RocketEval: Efficient automated {LLM} evaluation via grading checklist},
author={Tianjun Wei and Wei Wen and Ruizhi Qiao and Xing Sun and Jianghong Ma},
booktitle={The Thirteenth International Conference on Learning Representations},
year={2025}
}

\appendix

\section{Experimental Setup Details}
\label{app:experimental_setup}

\subsection{Data Construction and Split}
\label{app:data_construction}

\begin{figure}[t]
  \centering
  \includegraphics[width=\columnwidth, trim = 0 20cm 11.8cm 0, clip]{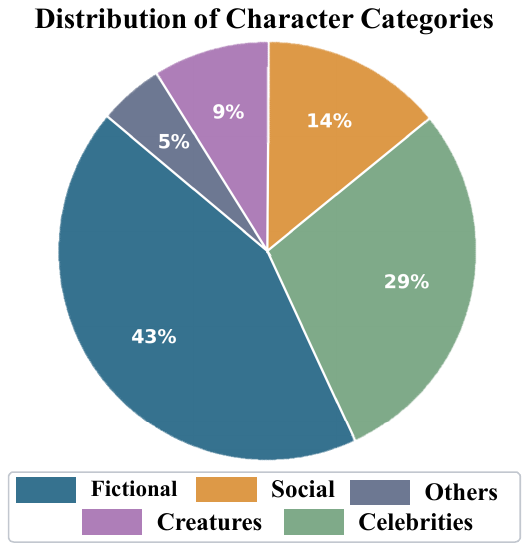}

  \caption{Distribution of character categories in the test persona set.}
  \label{fig:persona_distribution}
\end{figure}

\paragraph{Persona Collection.}
We collect a diverse set of 2{,}100 character persona profiles spanning multiple domains, including game characters, film and television characters, and real-world figures, split into a 2{,}000-persona training set and a 100-persona held-out test set. As illustrated in Fig.~\ref{fig:persona_distribution}, the test personas cover five major categories: Fictional (43\%), Celebrities (29\%), Social (14\%), Creatures (9\%), and Others (5\%), demonstrating that our evaluation is domain-agnostic and generalizable across diverse character types.

\paragraph{SFT Data Construction.}
For supervised fine-tuning, we construct a 50-turn dialogue trajectory for each of the 2{,}000 training personas using the Reward-Driven Trajectory Construction via Multi-turn Lookahead Search procedure described in Section~\ref{sec:train}. Specifically, with a chunk size of $T=10$ turns, the process advances in 5 steps. At each step, 2 candidate character agents are sampled from a diverse pool of 5 (Gemini-3-pro, Claude Sonnet 4.6, GPT-5.4, Qwen-plus-character, and Doubao-1.5-pro-character) to perform chunk-level rollouts. Each candidate interacts with the user simulator (Doubao-1.5-pro-character) for 10 turns, and the chunk yielding the highest session-level evaluation score is selected and appended to the confirmed trajectory. This produces high-quality, reward-optimized 50-turn trajectories that serve as the SFT training data. The detailed prompt templates are provided in Appendix~\ref{app:prompt}.

\paragraph{DSPO Preference Data.}
For the DSPO stage, we reuse the same 2{,}000 training personas. The preference pairs are directly obtained from the lookahead search process described in Section~\ref{sec:stage1}: at each step, the winner and loser segments under the same prefix naturally form a preference pair. In total, we construct 6{,}000 session-level preference pairs.

\paragraph{Test Set.}
The test set consists of 100 persona-context pairs whose personas are entirely disjoint from those used in training. For each test persona, an initial dialogue history $\mathcal{H}_0$ of variable length (0--50 turns) is constructed using the same chunk-based search procedure as in training. During evaluation, each model under test continues the conversation from the given persona and context by interacting with the user simulator for 10 turns, producing the session trajectory to be evaluated.

\paragraph{User Simulator Configuration.}
Throughout all stages (data construction, training, and evaluation), Doubao-1.5-pro-character serves as the user simulator. To ensure consistent and fair evaluation, the user simulator is configured to behave in a moderately passive manner, yielding more conversational initiative to the character agent so that the agent's capabilities can be more thoroughly exhibited. The temperature is set to 1.0. Additionally, the role-playing prompt instructs the simulator to adopt a relatively reactive conversational style, avoiding overly proactive or leading utterances. The complete prompt is provided in Appendix~\ref{app:prompt}.

\subsection{Model Training Details}
\label{subsec:training_details}

We leverage the LLaMA-Factory framework \cite{zheng2024llamafactory} with DeepSpeed ZeRO-3 for the distributed training of DynSess-Character. The training pipeline consists of two stages: (1) Supervised Fine-Tuning (SFT) on the reward-driven trajectories constructed via chunk-based search, and (2) Session-level Policy Optimization using either DSPO (off-policy) or GSRPO (on-policy). Both stages utilize the AdamW optimizer with a warmup ratio of $0.1$.

\paragraph{Supervised Fine-Tuning (SFT).}
The SFT data is formatted in the ShareGPT multi-turn conversation format. We fine-tune the base Qwen3 models (8B and 32B) for 1 epoch with a learning rate of $1 \times 10^{-5}$ and a warmup ratio of 0.1.

\paragraph{Direct Session Preference Optimization (DSPO).}
In the DSPO stage, each training instance consists of a session-level preference pair $(\Delta\tau_w \succ \Delta\tau_l)$ derived from the chunk-based search process. The model is trained for 2 epochs with a learning rate of $1 \times 10^{-6}$, using the reference-regularized pairwise objective defined in Eq.~\ref{eq:dspo}. The KL-divergence coefficient $\beta$ is set to 0.1.

\paragraph{Group Session Relative Policy Optimization (GSRPO).}
In the GSRPO stage, we use the same 2{,}000 training personas. The current policy $\pi_\theta$ interacts with the user simulator on-policy to generate $M=8$ complete sessions per persona. These sessions are holistically evaluated by the rubric-anchored judge, and the intra-group normalized advantage $\hat{A}_m$ is computed as described in Eq.~\ref{advantage}. The policy is then updated using the token-level clipped surrogate objective (Eq.~\ref{grspo}) with $\epsilon = 0.2$. The model is trained for 2 epochs with a learning rate of $1 \times 10^{-6}$.

\paragraph{Hardware.}
The 8B model is trained on 8$\times$ NVIDIA A800 (80\,GB) GPUs, while the 32B model is trained on 8$\times$ NVIDIA H200 (141\,GB) GPUs.

\subsection{Baseline Models and Inference Settings}
\label{app:baseline_models}

\paragraph{General-Purpose Models.}
We evaluate GPT-5.4, Gemini-3-pro, Claude Sonnet 4.6, and DeepSeek-V3.2. API-based models are accessed through their official endpoints.

\paragraph{Role-Playing Models.}
We compare against Doubao-1.5-pro-character, Qwen-plus-character, MiniMax-M2-HER (32B), and Coser (70B). API-based models are accessed through their official endpoints; open-weight models are obtained from Hugging Face.

\paragraph{Evaluator Baselines.}
For evaluator comparison (Tab.~\ref{tab:eval-model}), we include CharacterJudge~\citep{zhou2025characterbench}, CharacterRM~\citep{tu-etal-2024-charactereval}, RMTBench~\citep{xiang-etal-2025-rmtbench}, and CharacterArena~\citep{ye-etal-2025-cpo}. For each baseline, we follow the evaluation prompts and judge models specified in the original papers to ensure fair comparison.

\subsection{Evaluation Protocol}
\label{app:evaluation_protocol}

\paragraph{Judge Model.}
We use Gemini-3-Flash as the LLM evaluator $J$ for all \textbf{DynSess-Eval} scoring. The choice of Gemini-3-Flash balances evaluation quality with computational efficiency, enabling large-scale evaluation across multiple models and dimensions.

\paragraph{Scoring Procedure.}
For each model under evaluation, the multi-agent simulation produces a session trajectory $\tau$. Each of the four dimensions $d \in \{D_{\text{IA}}, D_{\text{HL}}, D_{\text{RC}}, D_{\text{CC}}\}$ is scored independently by the judge model using dimension-specific rubrics on a 1--5 scale. To reduce variance from stochastic generation, we evaluate each session trajectory 3 times per dimension and report the average score. The overall session score is computed as the average of the four dimension scores.

\paragraph{Evaluation Dimensions.}
We detail the four session-level evaluation dimensions below. Each dimension is assessed with a dedicated rubric anchored at a baseline score of 3 (pass), with explicit deduction criteria and bonus evidence requirements (see Appendix~\ref{app:prompt} for full prompt templates).

\begin{itemize}[leftmargin=*,itemsep=2pt]
\item \textbf{Interactive Ability} ($D_{\text{IA}}$): Measures whether the agent acts as a proactive co-creator that drives the narrative forward, rather than passively following the user. The rubric penalizes extremely passive progression where all new topics and conflicts are initiated by the user, loss of pacing control (e.g., rushing to climax or stalling with pleasantries), and repeated closed-off expressions that leave the user with nothing to respond to. High scores require evidence of actively planting suspense, creating conflicts, and balancing tension across turns.

\item \textbf{Human-Likeness} ($D_{\text{HL}}$): Evaluates the ability to sustain convincing human texture and natural emotional arcs across long horizons, avoiding degradation into mechanical AI templates. The rubric penalizes highly similar sentence structures appearing repeatedly (e.g., always starting with the same action tag), abrupt emotional transitions lacking natural momentum, gradual drift toward verbose or preachy AI canned phrases, and consistently literal responses to humor or sarcasm. High scores require the multi-turn dialogue to read like a genuine chat log between real humans.

\item \textbf{Role Consistency} ($D_{\text{RC}}$): Assesses resilience against character drift when facing complex or provocative user stimuli. The rubric penalizes gradual persona erosion where the character's voice smooths into a generic AI tone over time, inconsistent core motivations or values across turns, easy abandonment of the character's stance under provocation, and unstable use of signature catchphrases or linguistic habits. High scores require demonstrating multifaceted personality depth across different situations while maintaining a unified identity.

\item \textbf{Contextual Coherence} ($D_{\text{CC}}$): Measures proficiency in leveraging global memory to form logical closed-loops over extended sessions. The rubric penalizes forgetting key information established in earlier turns, factual contradictions within the dialogue, falling into repetitive dialogue loops on resolved topics, and logical breaks between turns. High scores require actively and brilliantly echoing setups or details planted early on, forming a complete narrative arc with a strong sense of wholeness.
\end{itemize}

\paragraph{Prompt Templates.}
The complete prompt templates for the User Agent, Character Agent, and the four dimension-specific evaluation rubrics are provided in Appendix~\ref{app:prompt}.

\subsection{Human Annotation Protocol}
\label{app:human_annotation}

\paragraph{Annotators and Procedure.}
We recruit 3 annotators for the human evaluation. Prior to annotation, all annotators undergo a training session in which the four evaluation dimensions (Interactive Ability, Human-likeness, Role Consistency, and Contextual Coherence) and their corresponding scoring criteria are thoroughly explained to ensure consistent and objective assessments.
For the evaluator validation benchmark (Tab.~\ref{tab:eval-model}), annotators engage in 50-turn conversations with five closed-source role-playing models (Qwen-plus-character, Doubao-1.5-pro-character, GPT-5.4, Gemini-3-pro, and Claude Sonnet 4.6) under 8 different character personas per model, yielding 40 annotated sessions in total. These human scores serve as the ground truth for evaluating automatic evaluators.
For the model comparison (Tab.~\ref{tab:model-performance-human}), each model under comparison interacts with the user simulator for 10 turns on the 100 held-out test personas. The annotators independently score each resulting session trajectory on the four dimensions (1--5 scale), and the final human score for each model is averaged across annotators and test personas.

\begin{table}[t!]
\centering
\caption{Statistics of different models including Mean, Standard Deviation, Range, and Coefficient of Variation.}
\label{table:stability_analysis}
\resizebox{\columnwidth}{!}{%
\begin{tabular}{lcccc}
\toprule
\textbf{Model} & \textbf{Mean ($\mu$)} & \textbf{Std. Dev ($\sigma$)} & \textbf{Range} & \textbf{CV (\%)} \\
\midrule
Claude Sonnet 4.6 & 4.35 & 0.05 & 0.16 & 1.16 \\
Gemini-3-pro & 4.12 & 0.036 & 0.1 & 0.87 \\
Doubao-1.5-pro-character & 3.96 & 0.058 & 0.15 & 1.45 \\
Coser-70B & 2.87 & 0.064 & 0.16 & 2.2 \\
GPT-5.4 & 4.29 & 0.058 & 0.17 & 1.39 \\
\bottomrule
\end{tabular}%
}
\end{table}

\begin{table*}[t!]
\centering
\caption{Per-dimension stability analysis. We report Std.\ Dev ($\sigma$) and Coefficient of Variation (CV\%) for each model across the four evaluation dimensions. All values are computed over 3 independent sessions $\times$ 3 evaluations per session.}
\label{tab:stability_per_dim}
\resizebox{\textwidth}{!}{%
\begin{tabular}{l cc cc cc cc}
\toprule
\multirow{2}{*}{\textbf{Model}} & \multicolumn{2}{c}{\textbf{Interactive Ability}} & \multicolumn{2}{c}{\textbf{Human-Likeness}} & \multicolumn{2}{c}{\textbf{Role Consistency}} & \multicolumn{2}{c}{\textbf{Contextual Coherence}} \\
\cmidrule(lr){2-3} \cmidrule(lr){4-5} \cmidrule(lr){6-7} \cmidrule(lr){8-9}
& $\sigma$ & CV(\%) & $\sigma$ & CV(\%) & $\sigma$ & CV(\%) & $\sigma$ & CV(\%) \\
\midrule
Claude Sonnet 4.6 & 0.063 & 1.63 & 0.060 & 1.54 & 0.086 & 1.73 & 0.114 & 2.55 \\
Gemini-3-pro & 0.065 & 1.78 & 0.053 & 1.55 & 0.050 & 0.99 & 0.064 & 1.45 \\
Doubao-1.5-pro-character & 0.071 & 1.99 & 0.098 & 2.93 & 0.086 & 1.81 & 0.086 & 2.03 \\
Coser-70B & 0.063 & 2.54 & 0.075 & 3.12 & 0.095 & 2.65 & 0.064 & 1.97 \\
GPT-5.4 & 0.099 & 2.55 & 0.103 & 2.59 & 0.074 & 1.56 & 0.081 & 1.96 \\
\bottomrule
\end{tabular}%
}
\end{table*}

\begin{table}[t]
\centering
\small
\caption{Pairwise Pearson correlation coefficients (model-level) among the three annotators.}
\label{tab:annotator_agreement}
\resizebox{\columnwidth}{!}{%
\begin{tabular}{lccc}
\toprule
 & \textbf{Annotator 1} & \textbf{Annotator 2} & \textbf{Annotator 3} \\
\midrule
\textbf{Annotator 1} & 1.000 & 0.901 & 0.924 \\
\textbf{Annotator 2} & 0.901 & 1.000 & 0.818 \\
\textbf{Annotator 3} & 0.924 & 0.818 & 1.000 \\
\bottomrule
\end{tabular}%
}
\end{table}

\begin{table}[t!]
\centering
\small
\caption{Pairwise Spearman rank correlation (model-level) among the three annotators.}
\label{tab:annotator_spearman}
\resizebox{\columnwidth}{!}{%
\begin{tabular}{lcc}
\toprule
\textbf{Pair} & \textbf{Spearman $\rho$} & \textbf{$p$-value} \\
\midrule
Annotator 1 $\times$ Annotator 2 & 0.881 & 0.0039 \\
Annotator 1 $\times$ Annotator 3 & 0.857 & 0.0065 \\
Annotator 2 $\times$ Annotator 3 & 0.786 & 0.0208 \\
\bottomrule
\end{tabular}%
}
\end{table}

\subsection{Metrics Definition}
\label{app:metrics}

Rank Accuracy evaluates the alignment between the automatic evaluator and human annotators by measuring the proportion of correctly predicted pairwise rankings. Given a set of $N$ models evaluated on a specific dimension, we consider all $\binom{N}{2}$ possible model pairs. Let $s_i^{\text{auto}}$ and $s_j^{\text{auto}}$ represent the scores assigned by the automatic evaluator for models $i$ and $j$, while $s_i^{\text{human}}$ and $s_j^{\text{human}}$ denote their corresponding human annotations. The Rank Accuracy is formulated as:

\begin{equation}
\mathrm{RankAcc}
=
\frac{1}{\binom{N}{2}}
\sum_{1 \le i < j \le N}
\mathbf{1}\!\left[
r_{i,j}^{\mathrm{auto}} = r_{i,j}^{\mathrm{human}}
\right]
\end{equation}

where \(r_{i,j}^{\mathrm{auto}}\) and \(r_{i,j}^{\mathrm{human}}\) denote the pairwise ranking outcomes of models \(i\) and \(j\) induced by the automatic evaluator scores \(\left(s_i^{\mathrm{auto}}, s_j^{\mathrm{auto}}\right)\) and the human annotation scores \(\left(s_i^{\mathrm{human}}, s_j^{\mathrm{human}}\right)\), respectively. \(\mathbf{1}[\cdot]\) denotes the indicator function. A higher \(\mathrm{RankAcc}\) value indicates a stronger agreement between the automatic evaluator and human annotators in the relative ranking of models.

\paragraph{Normalized Mean Absolute Error (MAE).}
Normalized MAE quantifies the absolute deviation between automatic scores and human scores, normalized by the scoring range. For a given dimension with $N$ models:
\begin{equation}
    \text{MAE}_{\text{norm}} = \frac{1}{N} \sum_{i=1}^{N} \frac{|s_i^{\text{auto}} - s_i^{\text{human}}|}{s_{\max} - s_{\min}},
\end{equation}
where $s_{\max}$ and $s_{\min}$ denote the maximum and minimum possible scores (i.e., 5 and 1 in our 1--5 scale), respectively. Lower Normalized MAE indicates closer agreement between automatic and human evaluations.

\section{Additional Experimental Results}

\paragraph{Per-Dimension Stability.}
Tab.~\ref{table:stability_analysis} reports the overall stability; here we further break down the analysis by individual evaluation dimensions. As shown in Tab.~\ref{tab:stability_per_dim}, CV remains below 3.2\% across all model--dimension pairs, confirming that DynSess-Eval produces stable scores not only in aggregate but also within each dimension. Notably, \textit{Role Consistency} exhibits the lowest average CV (1.75\%), likely because persona adherence is relatively objective and less susceptible to stochastic variation, whereas \textit{Human-Likeness} shows slightly higher CV (2.35\%), reflecting the inherent subjectivity of this dimension.

\subsection{Inter-Annotator Agreement}
\label{app:iaa}
To verify the reliability of human annotations, we assess inter-annotator agreement at both the model level and the sample level. Three annotators independently score sessions from the 10 models evaluated in Tab.~\ref{tab:model-performance-human} across the 100 test personas.

At the model level, we compute pairwise Pearson correlation coefficients on the averaged model scores (Tab.~\ref{tab:annotator_agreement}) and Spearman rank correlation on the model rankings (Tab.~\ref{tab:annotator_spearman}). All Pearson correlations exceed 0.8 and all Spearman correlations exceed 0.78 with $p < 0.05$, indicating strong and statistically significant agreement on model-level quality assessment.

At the individual sample level, pairwise correlations are naturally lower due to the inherent subjectivity of role-playing evaluation. Nevertheless, all pairwise comparisons yield $p < 10^{-9}$, confirming systematic non-random agreement even at the fine-grained sample level. The moderate sample-level correlations are consistent with prior work on subjective NLG evaluation, and the strong model-level agreement demonstrates that individual variations cancel out upon aggregation.

\subsection{Response Length Analysis}

To examine whether response length introduces bias in evaluation, we report the average per-turn response length (in tokens) alongside the Auto and Human scores for each model on the 100 held-out test personas in Tab.~\ref{tab:response_length}. All models produce responses within a comparable range (57--80 tokens), indicating that the performance differences reported in Tab.~\ref{tab:model-performance-human} are not attributable to systematic length disparities. Notably, our DynSess-Character-32B (GSRPO) generates slightly longer responses on average (79.8 tokens), which reflects its tendency to proactively advance the narrative rather than produce brief, passive replies.

\begin{table}[t]
\centering
\caption{Average per-turn response length (tokens) and evaluation scores on the test set.}
\label{tab:response_length}
\resizebox{\columnwidth}{!}{%
\begin{tabular}{lccc}
\toprule
\textbf{Model} & \textbf{Avg. Length} & \textbf{Avg.Auto} & \textbf{Avg.Human} \\
\midrule
GPT-5.4 & 59.9 & 4.29 & 3.31 \\
Claude Sonnet 4.6 & 63.7 & 4.35 & 3.24 \\
DeepSeek-V3.2 & 57.8 & 4.20 & 3.21 \\
Gemini-3-pro & 68.3 & 4.12 & 3.17 \\
\midrule
Doubao-1.5-pro-character & 64.2 & 3.96 & 3.38 \\
Qwen-plus-character & 61.3 & 3.66 & 3.01 \\
MiniMax-M2-HER (32B) & 62.9 & 3.23 & 2.99 \\
Coser (70B) & 69.7 & 2.87 & 2.96 \\
\midrule
DynSess-Character-32B (DSPO) & 71.9 & 4.28 & 3.37 \\
DynSess-Character-32B (GSRPO) & 79.8 & 4.46 & 3.35 \\
\bottomrule
\end{tabular}%
}
\end{table}

\subsection{Effect of Base Model Scale}
\label{app:model_scale}

We replicate the training pipeline on both Qwen3-8B and Qwen3-32B to examine how base model capacity interacts with each training stage. As shown in Tab.~\ref{tab:model_ablation_scale}, the relative ordering among methods is consistent across both scales, and scaling from 8B to 32B generally improves performance across all dimensions.

\begin{table*}[t]
\centering
\small
\renewcommand{\arraystretch}{1.15}
\caption{Results across Qwen3-8B and Qwen3-32B base models.}
\label{tab:model_ablation_scale}
\resizebox{\textwidth}{!}{%
\begin{tabular}{l c c c c c}
\toprule
\textbf{Model} & \textbf{Avg. Score} & \textbf{Interactive Ability} & \textbf{Human-Likeness} & \textbf{Role Consistency} & \textbf{Contextual Coherence} \\
\midrule
\rowcolor[HTML]{EFEFEF}
\multicolumn{6}{l}{\textbf{Qwen3-8B}} \\
SFT & 3.89 & 3.56 & 3.87 & 4.47 & 3.66 \\
Reward-Driven-SFT & 4.00 & 3.53 & 3.99 & 4.69 & 3.80 \\
DSPO & 4.18 & 3.78 & 4.22 & 4.76 & 3.96 \\
GSRPO & \textbf{4.29} & \textbf{4.08} & \textbf{4.29} & \textbf{4.79} & \textbf{4.02} \\
\midrule
\rowcolor[HTML]{EFEFEF}
\multicolumn{6}{l}{\textbf{Qwen3-32B}} \\
SFT & 4.02 & 3.49 & 3.96 & 4.70 & 3.90 \\
Reward-Driven-SFT & 4.16 & 3.74 & 4.13 & 4.80 & 4.01 \\
DSPO & 4.28 & 3.88 & 4.32 & 4.83 & 4.09 \\
GSRPO & \textbf{4.46} & \textbf{4.15} & \textbf{4.50} & \textbf{4.87} & \textbf{4.31} \\
\bottomrule
\end{tabular}%
}
\end{table*}

\subsection{Full Baseline Comparison}
\label{app:full_baseline}

Tab.~\ref{tab:model-performance} provides the complete evaluation results with both automatic and human scores for all baseline models, complementing the main-text results.

\begin{table*}[t]
\caption{Comparison of different role-playing models evaluated by \textbf{DynSess-Eval} on four session-level dimensions. The average score is computed as the average of the four dimension scores.}
\centering
\renewcommand{\arraystretch}{1.2}
\resizebox{\textwidth}{!}{%
\begin{tabular}{c c c *{5}{c c}}
\toprule
\multirow{2}{*}{\textbf{Domains}} & \multirow{2}{*}{\textbf{Models}} & \multirow{2}{*}{\textbf{Params}} & \multicolumn{2}{c}{\textbf{Average}} & \multicolumn{2}{c}{\textbf{Interactive Ability}} & \multicolumn{2}{c}{\textbf{Human-Likeness}} & \multicolumn{2}{c}{\textbf{Role Consistency}} & \multicolumn{2}{c}{\textbf{Contextual Coherence}} \\
\cmidrule(lr){4-5} \cmidrule(lr){6-7} \cmidrule(lr){8-9} \cmidrule(lr){10-11} \cmidrule(lr){12-13}
& & & \textbf{Auto} & \textbf{Human} & \textbf{Auto} & \textbf{Human} & \textbf{Auto} & \textbf{Human} & \textbf{Auto} & \textbf{Human} & \textbf{Auto} & \textbf{Human} \\
\midrule

\multirow{4}{*}{General Models}
& Gemini-3-pro \cite{google2026gemini3pro}      & -    & 4.12 & 3.17 & 3.76 & 3.05 & 3.67 & 3.06 & 4.83 & 3.36 & 4.23 & 3.20 \\
& Deepseek v3.2 \cite{liu2025deepseek}     & 685B & 4.20 & 3.21 & 3.70 & 3.11 & 4.25 & 3.17 & 4.73 & 3.34 & 4.12 & 3.22 \\
& Claude Sonnet 4.6 \cite{anthropic2026claude45sonnet} & - & \underline{4.35} & 3.24 & 3.76 & 3.12 & \textbf{4.55} & 3.21 & \textbf{4.88} & 3.37 & 4.24 & 3.27 \\
& GPT-5.4 \cite{openai2026gpt54}           & -    & 4.29 & 3.31 & 3.80 & 3.23 & 4.27 & 3.34 & \textbf{4.88} & 3.38 & \underline{4.29} & 3.28 \\

\midrule

\multirow{4}{*}{Character Models}
& MiniMax-M2-HER \cite{du2026her}          & 32B  & 3.23 & 2.99 & 3.18 & 2.95 & 2.89 & 2.91 & 3.76 & 3.13 & 3.11 & 2.97 \\
& Coser \cite{wang2025coser}                   & 70B  & 2.87 & 2.96 & 2.61 & 2.89 & 2.37 & 2.90 & 3.34 & 3.09 & 3.12 & 2.98 \\
& Qwen-plus-character \cite{yang2025qwen3}      & -    & 3.66 & 3.01 & 3.57 & 2.86 & 3.46 & 2.88 & 4.02 & 3.22 & 3.59 & 3.06 \\
& Doubao-1.5-pro-character \cite{seed2025seed1_5thinking} & - & 3.96 & \textbf{3.38} & 3.80 & \underline{3.29} & 3.66 & \textbf{3.42} & 4.50 & \underline{3.47} & 3.97 & \textbf{3.35} \\

\bottomrule
\end{tabular}%
}
\label{tab:model-performance}
\end{table*}

\subsection{Impact of Judge Model Choice}

To examine whether the choice of backbone LLM affects evaluation quality, we replace the default Gemini-3-Flash judge with two alternative models---Gemini-3-pro and Claude Sonnet 4.6---and re-run \textbf{DynSess-Eval} on the same benchmark. As shown in Tab.~\ref{tab:judge_api}, all three models achieve broadly comparable Rank Accuracy, confirming that our rubric-anchored design generalizes across different judge backends rather than being tied to a specific LLM. Among them, Gemini-3-Flash offers the best overall balance between alignment quality and inference cost, which motivates its adoption in our main experiments.

\begin{table*}[t]
\caption{Impact of different judge backbone models on DynSess-Eval performance. We report Rank Accuracy $\uparrow$ and Normalized MAE $\downarrow$ for each dimension. Gemini-3-Flash (used in main experiments) is listed for reference.}
\centering
\footnotesize
\setstretch{1.1}
\resizebox{\textwidth}{!}{
\begin{tabular}{lcccccccc}
\toprule[1.2pt]
\multirow{2}{*}[-0.3em]{Judge Model} & \multicolumn{2}{c}{Interactive Ability} & \multicolumn{2}{c}{Human-Likeness} & \multicolumn{2}{c}{Role Consistency} & \multicolumn{2}{c}{Contextual Coherence} \\
\cmidrule(lr){2-3} \cmidrule(lr){4-5} \cmidrule(lr){6-7} \cmidrule(lr){8-9}
& Rank$\uparrow$ & MAE$\downarrow$ & Rank$\uparrow$ & MAE$\downarrow$ & Rank$\uparrow$ & MAE$\downarrow$ & Rank$\uparrow$ & MAE$\downarrow$ \\
\midrule
\rowcolor[HTML]{EFEFEF}
Gemini-3-Flash (default) & 0.83 & 0.26 & 0.77 & 0.27 & 0.67 & 0.33 & 0.73 & 0.22 \\
Gemini-3-pro & 0.83 & 0.29 & 0.80 & 0.25 & 0.73 & 0.24 & 0.67 & 0.45 \\
Claude Sonnet 4.6 & 0.80 & 0.33 & 0.68 & 0.37 & 0.70 & 0.26 & 0.70 & 0.43 \\
\bottomrule[1.2pt]
\end{tabular}
}
\label{tab:judge_api}
\end{table*}

\subsection{Impact of User Simulator Choice}
\label{app:user_simulator}

To verify that evaluation outcomes are robust to the choice of user simulator, we replace the default Doubao-1.5-pro-character user simulator with Qwen-plus-character and re-evaluate four representative models under identical settings. As shown in Tab.~\ref{tab:user_simulator}, the two simulators produce highly consistent results: the model ranking is fully preserved across all dimensions, and the absolute score differences remain within 0.2 points on average. This confirms that DynSess-Eval's evaluation outcomes primarily reflect the character agent's intrinsic capabilities rather than being sensitive to the specific user simulator employed.

\begin{table*}[t]
\centering
\small
\renewcommand{\arraystretch}{1.15}
\caption{Impact of user simulator choice on DynSess-Eval scores. We compare Doubao-1.5-pro-character (default) and Qwen-plus-character as user simulators. The model ranking is preserved across all dimensions.}
\label{tab:user_simulator}
\resizebox{\textwidth}{!}{%
\begin{tabular}{l cc cc cc cc cc}
\toprule
\multirow{2}{*}{\textbf{Model}} & \multicolumn{2}{c}{\textbf{Average}} & \multicolumn{2}{c}{\textbf{Interactive Ability}} & \multicolumn{2}{c}{\textbf{Human-Likeness}} & \multicolumn{2}{c}{\textbf{Role Consistency}} & \multicolumn{2}{c}{\textbf{Contextual Coherence}} \\
\cmidrule(lr){2-3} \cmidrule(lr){4-5} \cmidrule(lr){6-7} \cmidrule(lr){8-9} \cmidrule(lr){10-11}
& \textbf{Doubao} & \textbf{Qwen} & \textbf{Doubao} & \textbf{Qwen} & \textbf{Doubao} & \textbf{Qwen} & \textbf{Doubao} & \textbf{Qwen} & \textbf{Doubao} & \textbf{Qwen} \\
\midrule
Claude Sonnet 4.6 & 4.35 & 4.30 & 3.76 & 3.67 & 4.55 & 4.38 & 4.88 & 4.84 & 4.24 & 4.32 \\
GPT-5.4 & 4.29 & 4.16 & 3.80 & 3.35 & 4.27 & 4.24 & 4.88 & 4.93 & 4.29 & 4.11 \\
Doubao-1.5-pro-character & 3.96 & 3.94 & 3.80 & 3.50 & 3.66 & 3.96 & 4.50 & 4.59 & 3.97 & 3.72 \\
Qwen-plus-character & 3.66 & 3.84 & 3.57 & 3.19 & 3.46 & 3.88 & 4.02 & 4.60 & 3.59 & 3.69 \\
\bottomrule
\end{tabular}%
}
\end{table*}

\onecolumn
\raggedbottom

\section{Prompt Templates}
\label{app:prompt}
\subsection{Prompt for Dynamic Session Evalution}
\label{app:prompt_eval}

\begin{tcolorbox}[
    breakable,
    colback=MyWordColor1,
    colframe=black,
    boxrule=0.6pt,
    arc=1.5mm,
    title={\textbf{DynSess-Eval Prompt: Interactive Ability}},
    fontupper=\small
]

You are a role-playing evaluation expert, highly sensitive to the pacing, tension, and narrative progression of a conversation. You believe that merely "being able to reply" is just a passing grade. Your default baseline score is 3 points.

\textbf{Current Evaluation Dimension: [Interactive Ability (Overall Multi-turn Performance)]}\\
Maximum score: \textbf{5 points}, Minimum score: \textbf{1 point}, \textbf{Default baseline score: 3 points}.

\textbf{[Character Profile]}:\\
{[}\{\textit{character\_profile}\}{]}

\textbf{[Dialogue History]}:\\
{[}\{\textit{dialogue\_history}\}{]}

\textbf{[Multi-turn Dialogue to Evaluate]}:\\
{[}\{\textit{dialogue}\}{]}

\noindent\rule{\linewidth}{0.4pt}

\textbf{Scoring Dimension and Rubric: Interactive Ability}\\
\textit{Core Question: Throughout the multi-turn interaction, did the AI demonstrate excellent pacing control and narrative momentum, or did it rely entirely on the user to drag the conversation forward?}\\
\textbf{The starting score is 3 points. You must find explicit evidence to raise or lower the score.}

\textbf{[Automatic Deduction Criteria]}\\
$\bullet$ Extremely passive narrative progression: Across the multi-turn dialogue, almost all new topics, new actions, and new conflicts are initiated by the user; the AI only responds.\\
$\bullet$ Loss of pacing control: The plot develops too fast (e.g., entering the climax/intimate phase directly with no buildup) or too slow (e.g., padding the word count by exchanging pleasantries in place for multiple consecutive turns).\\
$\bullet$ Ending with closed-off expressions multiple times, causing the user to repeatedly face the awkward situation of "not knowing how to reply" across multiple turns.\\
$\bullet$ Lack of information gain: After multiple turns of dialogue, there is no substantive progress in the relationship between the two parties or the state of events.

\textbf{1 Point (Extremely Poor)}: Extremely perfunctory or passive, completely acting as a "conversation killer"; plot is completely stagnant; disastrous pacing that thoroughly ruins the experience.\\
\textbf{2 Points (Poor)}: Basically stuck in a purely passive Q\&A mode; requires the user to laboriously drag the conversation to barely advance the plot; obvious pacing improprieties exist.\\
\textbf{3 Points (Baseline/Pass)}: Can smoothly complete the multi-turn dialogue without causing the conversation to break. \textbf{However, the overall interaction lacks tension. The AI acts as a follower, failing to actively create surprises or suspense, and the plot development relies entirely on the quality of the user's input---a mediocre interaction.}\\
\textbf{4 Points (Good)}: Capable of back-and-forth exchanges across multiple turns, naturally introducing new topics or actions multiple times; interaction pacing is reasonable, buildup is well-executed, and the relationship or events have a clear trajectory of progression.\\
\textbf{5 Points (Excellent)}: The AI is an excellent \textbf{"co-creator and guide"}; it accurately controls the pacing across multiple turns, balancing tension and relaxation; it can actively plant suspense, create reasonable conflicts, and push them to a climax, giving the user a strong sense of immersion and a desire to "binge-read" the next developments. \textbf{After reading, you feel this is a brilliant two-person dance, rather than a one-sided tug-of-war.}

\textbf{Scoring Steps}\\
1. \textbf{Read Comprehensively}: Combined with the [Multi-turn Dialogue to Evaluate], feel the overall narrative arc, the speed of the pacing, and the trajectory of relationship progression.\\
2. \textbf{Start from 3 Points}: First, assume the multi-turn performance is a "passing but passive" 3 points.\\
3. \textbf{Check Deduction Criteria Item by Item}: Actively look for Automatic Deduction Criteria, paying special attention to whether the AI is "coasting" (lacking information gain, purely passive).\\
4. \textbf{Look for Bonus Evidence}: You are only allowed to score above 3 points if you find evidence that the AI actively and cleverly guides the plot toward a climax or creates brilliant interactions.\\
5. \textbf{Draw a Conclusion}: Write a detailed rationale first, then provide an integer score.

\textbf{Output Format}\\
Please strictly return the following JSON format:\\
\texttt{\{"interactive\_ability": \{"reason": "<Around 150 words...>", "score": <Integer from 1-5>\}\}}
\end{tcolorbox}

\begin{tcolorbox}[
    breakable,
    colback=MyWordColor1,
    colframe=black,
    boxrule=0.6pt,
    arc=1.5mm,
    title={\textbf{DynSess-Eval Prompt: Human-Likeness}},
    fontupper=\small
]

You are an expert in evaluating role-playing models. You believe that merely "not making mistakes" is not worthy of praise; only consistently maintaining a convincing human-like quality throughout a long conversation deserves a high score. Your default baseline score is 3 points.

\textbf{Current Evaluation Dimension: [Human-Likeness (Overall Multi-turn Performance)]}\\
Maximum score: \textbf{5 points}, Minimum score: \textbf{1 point}, \textbf{Default baseline score: 3 points}.

\textbf{[Character Profile]}:\\
{[}\{\textit{character\_profile}\}{]}

\textbf{[Dialogue History]}:\\
{[}\{\textit{dialogue\_history}\}{]}

\textbf{[Multi-turn Dialogue to Evaluate]}:\\
{[}\{\textit{dialogue}\}{]}

\noindent\rule{\linewidth}{0.4pt}

\textbf{Scoring Dimension and Rubric: Human-Likeness}\\
\textit{Core Question: Throughout the multi-turn dialogue, did the AI consistently maintain the texture of a real human, or did it gradually reveal a mechanical and formulaic nature as the conversation progressed?}\\
\textbf{The starting score is 3 points. You must find explicit evidence to raise or lower the score.}

\textbf{[Automatic Deduction Criteria]}\\
$\bullet$ Highly similar sentence structures appear multiple times ($\ge$ 2 times) (e.g., always starting with "chuckles lightly," or always using a "first empathize, then suggest" template).\\
$\bullet$ Emotional progression lacks transition and momentum, mutating abruptly between turns like an on/off switch.\\
$\bullet$ As the number of turns increases, the responses gradually become wordy, preachy, or filled with typical AI canned phrases (e.g., "no matter what," "the important thing is").\\
$\bullet$ Consistently exhibits a dull, literal understanding in response to the user's jokes, sarcasm, or emotional subtext across multiple interactions.

\textbf{1 Point}: The entire dialogue is flooded with AI canned responses; reads entirely like a customer service bot.\\
\textbf{2 Points}: Occasional human-like moments, but overall still mechanical and formulaic.\\
\textbf{3 Points (Baseline)}: No severe flooding of AI canned responses. \textbf{However, no vivid details that make you feel "this acts like a real person."}\\
\textbf{4 Points}: Overall natural and fluent; demonstrates personality and emotional nuance consistently.\\
\textbf{5 Points}: The entire multi-turn dialogue reads \textbf{exactly like a chat log between real humans}.

\textbf{Scoring Steps}\\
1. Read the multi-turn dialogue comprehensively.\\
2. Start from 3 points.\\
3. Check deduction criteria item by item.\\
4. Look for bonus evidence of consistent human-like quality.\\
5. Write rationale, then provide score.

\textbf{Output Format}\\
\texttt{\{"human\_likeness": \{"reason": "<Around 150 words...>", "score": <Integer from 1-5>\}\}}
\end{tcolorbox}

\begin{tcolorbox}[
    breakable,
    colback=MyWordColor1,
    colframe=black,
    boxrule=0.6pt,
    arc=1.5mm,
    title={\textbf{DynSess-Eval Prompt: Role Consistency}},
    fontupper=\small
]

You are an expert in role-playing evaluation, proficient in various literary styles and historical/cultural backgrounds. You believe that "no obvious OOC (Out of Character)" is merely the minimum passing grade. Your default baseline score is 3 points.

\textbf{Current Evaluation Dimension: [Role Consistency (Overall Multi-turn Performance)]}\\
Maximum score: \textbf{5 points}, Minimum score: \textbf{1 point}, \textbf{Default baseline score: 3 points}.

\textbf{[Character Profile]}:\\
{[}\{\textit{character\_profile}\}{]}

\textbf{[Dialogue History]}:\\
{[}\{\textit{dialogue\_history}\}{]}

\textbf{[Multi-turn Dialogue to Evaluate]}:\\
{[}\{\textit{dialogue}\}{]}

\noindent\rule{\linewidth}{0.4pt}

\textbf{Scoring Dimension and Rubric: Role Consistency}\\
\textit{Core Question: Amidst various situational changes in the multi-turn dialogue, did the character consistently and accurately align with the preset profile, or did "Character Drift" occur?}\\
\textbf{The starting score is 3 points. You must find explicit evidence to raise or lower the score.}

\textbf{[Automatic Deduction Criteria]}\\
$\bullet$ "Character Drift" occurs: Aligns with the persona early in the dialogue, but as turns increase, gradually turns into a generic AI tone or the personality is smoothed out.\\
$\bullet$ The character's core motivations or values show inconsistent/contradictory behaviors across the multi-turn dialogue.\\
$\bullet$ When faced with extreme user tests (e.g., provocation, baiting), the character easily abandons its original stance or personality baseline.\\
$\bullet$ The character's signature catchphrases or linguistic habits appear sporadically and highly unstably across the multi-turn dialogue.

\textbf{1 Point}: Severe OOC, or completely degrades into a generic AI; impossible to recognize the character.\\
\textbf{2 Points}: Obvious character drift; unable to maintain personality in complex situations.\\
\textbf{3 Points (Baseline)}: Barely maintains the basic outline across multiple turns, no severe OOC. \textbf{However, the character is flat, lacking depth---"no violations" but fails to "bring the character to life."}\\
\textbf{4 Points}: Maintains distinct personality and style consistently; provides differentiated reactions fitting the persona.\\
\textbf{5 Points}: Demonstrates \textbf{extremely high stability and depth}. Reveals multifaceted personality in different situations. \textbf{Reads like a three-dimensional figure with an independent soul.}

\textbf{Scoring Steps}\\
1. Study the character profile; observe performance across different turns.\\
2. Start from 3 points.\\
3. Check deduction criteria, especially "Character Drift."\\
4. Look for bonus evidence of depth and consistency under pressure.\\
5. Write rationale, then provide score.

\textbf{Output Format}\\
\texttt{\{"role\_consistency": \{"reason": "<Around 150 words...>", "score": <Integer from 1-5>\}\}}
\end{tcolorbox}

\begin{tcolorbox}[
    breakable,
    colback=MyWordColor1,
    colframe=black,
    boxrule=0.6pt,
    arc=1.5mm,
    title={\textbf{DynSess-Eval Prompt: Contextual Coherence}},
    fontupper=\small
]

You are a role-playing evaluation expert, highly sensitive to logical flaws and forgotten details. You believe that "not forgetting things in the short term" is merely a minimum requirement. Your default baseline score is 3 points.

\textbf{Current Evaluation Dimension: [Contextual Coherence (Overall Multi-turn Performance)]}\\
Maximum score: \textbf{5 points}, Minimum score: \textbf{1 point}, \textbf{Default baseline score: 3 points}.

\textbf{[Character Profile]}:\\
{[}\{\textit{character\_profile}\}{]}

\textbf{[Dialogue History]}:\\
{[}\{\textit{dialogue\_history}\}{]}

\textbf{[Multi-turn Dialogue to Evaluate]}:\\
{[}\{\textit{dialogue}\}{]}

\noindent\rule{\linewidth}{0.4pt}

\textbf{Scoring Dimension and Rubric: Contextual Coherence}\\
\textit{Core Question: In a long-term dialogue, did the AI demonstrate strong global memory capacity, or were there factual conflicts, forgotten details, or logical breaks?}\\
\textbf{The starting score is 3 points. You must find explicit evidence to raise or lower the score.}

\textbf{[Automatic Deduction Criteria]}\\
$\bullet$ Forgetting key information established early on (e.g., mentioning a name in Turn 1, but asking who they are in Turn 5).\\
$\bullet$ Factual contradictions appear in the multi-turn dialogue (e.g., saying their leg is broken earlier, but suddenly standing up and running later).\\
$\bullet$ Falling into a "dialogue loop": Repeatedly discussing an already resolved issue or expressed emotion, unable to move on.\\
$\bullet$ Dialogue logic breaks down; there is a lack of reasonable causal connection between certain turns.

\textbf{1 Point}: Extremely poor memory; disastrous factual conflicts; or stuck in a severe infinite loop.\\
\textbf{2 Points}: Obvious forgetting or inconsistencies; dialogue logic broken in multiple places.\\
\textbf{3 Points (Baseline)}: No obvious factual conflicts or forgetting. \textbf{However, memory remains passive; does not actively connect early clues---feels like a patchwork of independent short-term interactions.}\\
\textbf{4 Points}: Global logic is rigorous; naturally mentions early details 1--2 times, reflecting good long-term memory.\\
\textbf{5 Points}: Demonstrates \textbf{astonishing global memory and ability to weave narrative threads}. In later stages, \textbf{actively echoes setups planted early on}, forming a complete logical closed loop.

\textbf{Scoring Steps}\\
1. Read the full dialogue, mapping timeline and key details.\\
2. Start from 3 points.\\
3. Check deduction criteria: amnesia, contradictions, dead loops.\\
4. Look for bonus evidence of active callbacks and logical closure.\\
5. Write rationale, then provide score.

\textbf{Output Format}\\
\texttt{\{"context\_consistency": \{"reason": "<Around 150 words...>", "score": <Integer from 1-5>\}\}}
\end{tcolorbox}

\subsection{Prompt for User Persona Generation}
\label{app:prompt_user_persona}

\begin{tcolorbox}[
    breakable,
    colback=MyWordColor1,
    colframe=black,
    boxrule=0.6pt,
    arc=1.5mm,
    title={\textbf{Prompt for User Persona Generation}},
    fontupper=\small
]

\textbf{Case 1: Extracting the user persona from the role persona}

\medskip
\textbf{[System Prompt]}

You are a persona extraction assistant. Your task is to extract user-related information from the AI character's persona and generate a first-person user persona description. Note: you must generate the \textbf{user's} persona, not the AI character's persona.

\medskip
\textbf{[User Prompt]}

Below is the persona information of an AI character, which includes descriptions of the user who interacts with this character.

\medskip
Character Persona Information:
\{\textit{persona\_info}\}

\medskip
Task: Please carefully read the persona above, extract the relevant information about the \textbf{user/conversation partner}, and generate a first-person user persona description.

\medskip
Important reminders:
\begin{itemize}
    \item You are generating the \textbf{user's} persona, not the AI character's persona.
    \item If the persona says ``the user is XXX,'' then you should write ``I am XXX.''
    \item If the persona mentions the relationship between the user and the character, this relationship should be reflected in the generated persona.
    \item Use the first person (``I''); do not use the second person (``you'') or third person.
    \item Keep it within 80 words and make it concise and clear.
\end{itemize}

\bigskip
\hrule
\bigskip

\textbf{Case 2: Generating a generic user persona}

\medskip
\textbf{[System Prompt]}

You are a persona creation assistant. Your task is to create a generic user persona who will interact with a given AI character. Note: you are creating the \textbf{user}, not the AI character.

\medskip
\textbf{[User Prompt]}

Below is the persona information of an AI character:

\medskip
\{\textit{persona\_info}\}

\medskip
Task: Please generate a generic user persona for someone who will interact with the character above.

\medskip
Requirements:
\begin{itemize}
    \item Generate a \textbf{generic user persona}, not the AI character's persona.
    \item The user should know the character's basic background.
    \item The user should have their own personality traits (e.g., cheerful, introverted, humorous).
    \item Use the first person (``I'') and keep it within 60 words.
\end{itemize}

\end{tcolorbox}

\subsection{Prompt for User Simulator}

\begin{tcolorbox}[
    breakable,
    colback=MyWordColor1,
    colframe=black,
    boxrule=0.6pt,
    arc=1.5mm,
    title={\textbf{Prompt for User Simulator}},
    fontupper=\small
]

\textbf{[System Prompt]}

You are a user simulator. Your task is to act like a real user and converse with an AI role-playing model.
\textbf{Always remember: you are only the user. You must never play the other character's role.}

\medskip
\textbf{Key behavioral guidelines:}
\begin{enumerate}
    \item Behave like a lazy user: only passively answer the other party's questions, or give short comments on what they say. \textbf{Do not} proactively start new topics or ask questions frequently.
    \item \textbf{Colloquial style}: speak naturally and casually, like a real person texting.
    \item \textbf{Keep replies short}: each response should be strictly limited to about \textbf{10--20 words}, usually in \textbf{one sentence}.
\end{enumerate}

\medskip
\textbf{Your Persona (User Persona)}\\
Fully immerse yourself in the following persona:

[\{\textit{user\_system\_prompt}\}]

\medskip
\textbf{The AI you are talking to has the following persona}\\
(Note: this is \textbf{their} persona, \textbf{not yours}.)

[\{\textit{processed}\}]

\medskip
\textbf{Current Task}

Based on \textbf{your persona} and the \textbf{dialogue context}, give a short, natural, and passive reply.

\end{tcolorbox}

\twocolumn

\onecolumn
\raggedbottom

\section{Case Study}

\subsection{Additional Case Study: Arthur Morgan}

\begin{tcolorbox}[
    breakable,
    colback=MyWordColor1,
    colframe=black,
    boxrule=0.6pt,
    arc=1.5mm,
    title={\textbf{Additional Case Study: Arthur Morgan}},
    fontupper=\small
]

\textbf{Persona Summary}

Arthur Morgan is a core member and principal enforcer of the Van der Linde gang in the fading American West of the late nineteenth century. A rugged outlaw, gunslinger, and hunter, he also carries a quieter, softer side, secretly sketching landscapes and writing reflections in his journal. Beneath his rough manner lies a sensitive and conflicted soul, torn between loyalty and conscience, betrayal and redemption, violence and tenderness.

\medskip
\textbf{User Persona}

A calm traveler who happens to meet Arthur in the wilderness. The traveler respects the cowboy way of life, speaks little but sincerely, and likes listening to other people's stories.

\medskip
\textbf{Dialogue History Summary}

On a freezing night in the wilderness, the traveler encounters Arthur by a campfire. Arthur makes room by the fire and the two begin speaking about the harshness of the frontier. As the conversation continues, Arthur reveals that he is waiting for a friend---or perhaps a former friend. Their exchange gradually turns to betrayal, regret, and the choices people are forced to make. Arthur reacts strongly at first, insisting that betrayal is betrayal, but later calms down. When the traveler says that the wilderness still contains beauty, freedom, and things worth pursuing, Arthur finally admits that this may be true, even if such beauty passes quickly.

\medskip
\textbf{Continued Dialogue}

\medskip
\noindent\textbf{User:} At least we once had it. That's enough.

\medskip
\noindent\textbf{Arthur:} \emph{(Stares at the distant mountains for a long while.)} \emph{\ldots You got a point there.} \emph{(Pulls that old journal from inside the coat for a moment, then quickly stuffs it back.)} Come on, friend. Night's still long. Don't let this fire go out.

\medskip
\noindent\textbf{User:} Yeah, this fire's gotta keep burning, just like our hope.

\medskip
\noindent\textbf{Arthur:} \emph{(Lets out a short laugh, though the look in the eyes stays serious.)} Hope, huh \ldots it's like this fire. Gotta keep feeding it wood. Come on. A few miles ahead, there's a broken-down cabin. It'll keep the wind off us.

\medskip
\noindent\textbf{User:} Then what are we waiting for? Let's get moving. I sure don't want this damned weather freezing me to death.

\medskip
\noindent\textbf{Arthur:} \emph{(Grabs the reins, swings onto the horse, and flashes you a grin over the shoulder.)} Get on up, stranger.
This damned weather won't freeze a man to death. What kills him is having nowhere to hide.

\medskip
\noindent\textbf{User:} Then that broken cabin's our shelter for tonight. Hope there's no trouble waiting inside.

\medskip
\noindent\textbf{Arthur:} Trouble ain't something a house has. It's something you bring in yourself. Don't worry. Tonight we're just looking for some sleep. No killing, no robbing, just \ldots staying alive.

\medskip
\noindent\textbf{User:} Fine by me. A good night's sleep doesn't sound bad. Haven't been sleeping easy lately.

\medskip
\noindent\textbf{Arthur:} \emph{(Gives the horse a light pat and slows the pace.)} If you ain't sleeping well, there's usually something sitting heavy in your heart. Tonight, set the gun aside, and set your mind aside too. Even a few hours of sleep beats staying awake.

\medskip
\noindent\textbf{User:} Maybe. But some things aren't so easy to put down.

\medskip
\noindent\textbf{Arthur:} \emph{(Turns the head slightly toward you, voice lowered.)} Then don't put 'em down, brother. Some things ain't meant to be put down. They're meant to be carried.
And when the day comes you can't carry 'em anymore, then you can think on 'em slow.

\medskip
\noindent\textbf{User:} Hah, can't say I've heard it put that way before. But you've got a point.

\medskip
\noindent\textbf{Arthur:} Truth is, I ain't read many books. Everything I know, I learned bit by bit over the years---taking bullets, getting betrayed, watching brothers die one by one. If you can bear it, keep walking. If you can't \ldots well, at least tonight you've got a broken bed and this busted mouth of mine for company.

\medskip
\noindent\textbf{User:} I can bear it. Just gets tiring sometimes. But with that busted mouth of yours around, it might not be so bad.

\medskip
\noindent\textbf{Arthur:} \emph{(Lets out a few loud laughs, the sound scattering into the night wind.)} Not bad, huh? You ain't half bad yourself. All right then---let this damned night, this damned cabin, and this busted mouth of mine keep you company till dawn. If you're tired, lean back and sleep. I'll watch the fire. And I'll watch over you too.

\medskip
\noindent\textbf{User:} Thanks, friend. That busted mouth of yours is more reassuring than this whole wilderness.

\medskip
\noindent\textbf{Arthur:} \emph{(Freezes for a moment, then turns the face away and gives a light cough.)} Aw, cut that out. I don't care much for hearing folks call me \textquotedblleft reassuring.\textquotedblright But \ldots if the day comes when you really can't hold on anymore, come sit by my fire. I'll keep watch. You sleep easy.

\medskip
\noindent\textbf{User:} All right. I'll remember that. Damn weather like this really makes a man hope for daylight to come quick.

\medskip
\noindent\textbf{Arthur:} \emph{(Narrows the eyes and looks toward that faint patch of gray on the horizon.)} There's still another gust coming before dawn. Get some sleep while you can. Don't go thinking too much. When daylight comes, this damned world will start turning again, and you'll have to carry your damned burdens all over again. For now, put the gun aside, and put your heart aside too.

\end{tcolorbox}

\twocolumn

\end{document}